\documentclass[10pt,twocolumn,letterpaper]{article}

 \usepackage{cvpr}              

\renewcommand{\paragraph}[1]{\vspace{.5em}\noindent\textbf{#1.}}

\newtheorem{definition}{Definition}

\definecolor{awesome}{rgb}{1.0, 0.13, 0.32}
\definecolor{azure}{rgb}{0.0, 0.5, 1.0}
\definecolor{amber}{rgb}{1.0, 0.75, 0.0}
\definecolor{deepmagenta}{rgb}{0.8, 0.0, 0.8}
\definecolor{darkpastelgreen}{rgb}{0.01, 0.75, 0.24}
\definecolor{deepskyblue}{rgb}{0.0, 0.75, 1.0}
\definecolor{deepcarminepink}{rgb}{0.94, 0.19, 0.22}

\newcommand{\bigparallel}{\mathrel{\|}}

\definecolor{cvprblue}{rgb}{0.21,0.49,0.74}
\usepackage[pagebackref,breaklinks,colorlinks,allcolors=cvprblue]{hyperref}

\usepackage{amsmath}
\usepackage{amssymb}
\usepackage{mathtools}
\usepackage{bm}
\usepackage{booktabs}
\usepackage{multirow}
\usepackage{xcolor}
\usepackage{microtype}
\usepackage{siunitx}

\usepackage{tikz}
\usetikzlibrary{calc}

\usepackage{algorithm}
\usepackage[noend]{algpseudocode}

\usetikzlibrary{arrows,shapes,positioning, decorations, babel, quotes}
\usetikzlibrary{arrows.meta}

\usepackage[babel,german=quotes]{csquotes}

\usepackage{pgfplots, pgfplotstable}

\usepackage{ltablex}

\usepackage{adjustbox}

\usepackage{orcidlink}

\def\paperID{*****} 
\def\confName{CVPR}
\def\confYear{2025}

\title{Graph Collaborative Attention Network for Link Prediction in Knowledge Graphs}

\author{\textbf{Thanh Hoang-Minh \orcidlink{0009-0007-0898-5923}}\\
Department of Information Technology, VNUHCM - University of Science\\
Ho Chi Minh City, Vietnam \\
{\tt\small hmthanhgm@gmail.com}
}

\begin{document}
\maketitle
\begin{abstract}
Knowledge graphs offer a structured representation of real-world entities and their relationships, enabling a wide range of applications from information retrieval to automated reasoning. In this paper, we conduct a systematic comparison between traditional rule-based approaches and modern deep learning methods for link prediction. We focus on KBGAT, a graph neural network model that leverages multi-head attention to jointly encode both entity and relation features within local neighborhood structures. To advance this line of research, we introduce \textbf{GCAT} (Graph Collaborative Attention Network), a refined model that enhances context aggregation and interaction between heterogeneous nodes. Experimental results on four widely-used benchmark datasets demonstrate that GCAT not only consistently outperforms rule-based methods but also achieves competitive or superior performance compared to existing neural embedding models. Our findings highlight the advantages of attention-based architectures in capturing complex relational patterns for knowledge graph completion tasks.
\end{abstract}
    
\section{Introduction}
\label{sec:intro}

Nowadays, graphs have been applied in all aspects of life. Social network graphs (e.g., Facebook \cite{ugander2011anatomy}) illustrate how individuals are connected to each other, the places we visit, and the information we interact with. Graphs are also used as core structures in video recommendation systems (e.g., YouTube \cite{baluja2008video}), flight networks, GPS navigation systems, scientific computations, and even brain connectivity analysis. Google’s Knowledge Graph \cite{googlekg:2020}, introduced in 2012 \cite{ji2020survey}, is a notable example of how information can be structured and utilized in knowledge graphs.

Effectively exploiting knowledge graphs provides users with deeper insight into the underlying data, which can benefit many real-world applications. However, in practice, new knowledge is continuously generated, and the acquired information is often incomplete or missing. This leads to the problem of knowledge graph completion or link prediction in knowledge graphs.

Most current approaches aim to predict a new edge connecting two existing nodes. Such methods help make the graph more complete—i.e., denser—by introducing additional connecting edges. However, these approaches primarily address the problem of completion rather than the challenge of integrating new knowledge into the graph, which remains an open question. Currently, research in knowledge graph completion follows two main directions: one is optimizing an objective function to make predictions with minimal error, as in RuDiK \cite{ortona2018robust}, AMIE \cite{galarraga2015fast}, and RuleN \cite{meilicke2018fine}, which are typically used in vertex or edge classification applications. The other approach generates a ranked list of \(k\) candidate triples, where the score reflects decreasing confidence, as seen in studies such as TransE \cite{bordes2013translating} and ConvKB \cite{vu2019capsule}, which are commonly used in recommendation systems. Our approach follows this second direction of producing a candidate list.

Within these approaches, there are two main methodologies: rule-based systems and embedding-based methods such as ConvE \cite{dettmers2017convolutional}, TransE \cite{bordes2013translating}, and ComplEx \cite{trouillon2016complex}. With the goal of gaining a systematic understanding of these methods, we chose to explore the embedding-based approach, and in particular, we selected KBAT \cite{nathani2019learning}, which employs attention mechanisms.

For the embedding-based method, we present a review of attention mechanisms \cite{vaswani2017attention}, their application in knowledge graphs via Graph Attention Networks (GATs) \cite{velivckovic2017graph}, and the KBAT model \cite{nathani2019learning}. Our contribution in the deep learning approach includes a publicly accessible implementation and training process on the project homepage\footnote{\url{https://graphattentionnetwork.github.io/}}, with both the training code and model results openly provided.

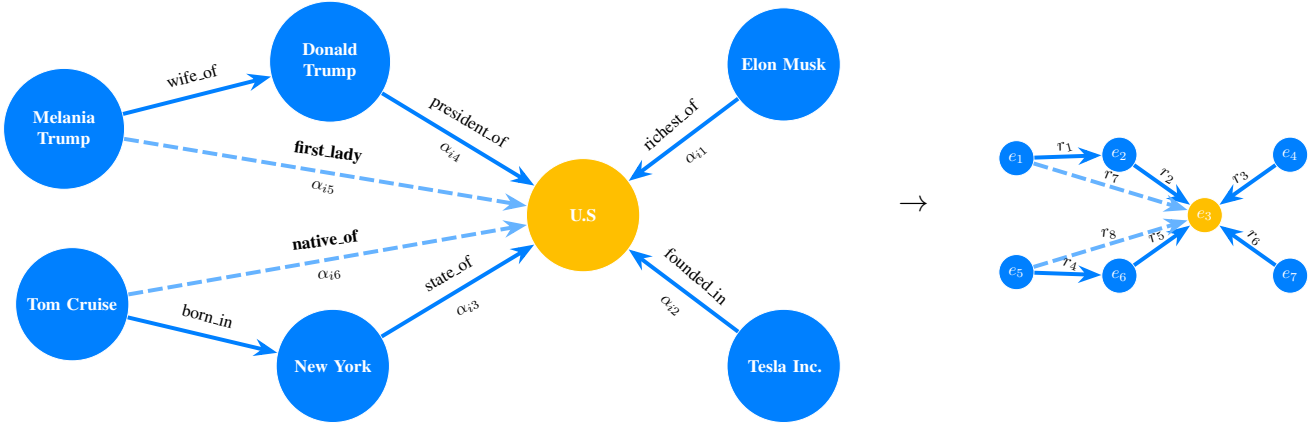
\begin{figure*}[htp]
	\centering
	\resizebox{\textwidth}{!}{%
		\begin{tikzpicture}[>=Stealth,
	entityStyle/.style={draw, color=white, thin, align=center, circle, text width=20mm},
	us/.style = {entityStyle, fill=amber},
	entity/.style={entityStyle, fill=azure},
	arrowStyle/.style={-latex', >=stealth },
	entityId/.style={draw, color=white,  thin, align=center, circle},]
	
	\node[us] (us) {\textbf{U.S}};
	
	\begin{scope}[every node/.style={entity}]
		\node[above left=2cm of us, xshift=-20mm] (c1) {\textbf{Donald Trump}};
		\node[above right=2cm of us, xshift=1cm] (c2) {\textbf{Elon Musk}};
		\node[below left=2cm of us, xshift=-20mm] (c3) {\textbf{New York}};
		\node[below right=2cm of us, xshift=1cm] (c4) {\textbf{Tesla Inc.}};
		
		\node[below left=3cm of c1, yshift=25mm, xshift=-15mm] (c1x) {\textbf{Melania Trump}};
		\node[above left=3cm of c3, yshift=-25mm, xshift=-15mm] (c3x) {\textbf{Tom Cruise}};
		
	\end{scope}
	
	
	\foreach \value in {1,...,4}
	\draw[->, line width=0.8mm, azure] (c\value) -> (us);
	\foreach \value in {1,3}
	\draw[->, line width=0.8mm, azure] (c\value x) -> (c\value);
	
	\foreach \value in {1,3}
	\draw[->, line width=0.8mm, azure!60, dash pattern=on 3mm off 1mm, postaction={decorate}] (c\value x) -> (us);
	
	\path[arrowStyle] (c1)--node[rotate=-31, yshift=4mm]{president\_of} node[rotate=-31, yshift=-3mm]{$\alpha_{i4}$} (us);
	
	\path[arrowStyle] (c3)--node[rotate=30, yshift=4mm]{state\_of} node[rotate=30, yshift=-4mm]{$\alpha_{i3}$} (us);
	
	\path[arrowStyle] (c2)--node[rotate=41, yshift=4mm]{richest\_of} node[rotate=41, yshift=-4mm]{$\alpha_{i1}$} (us);
	
	\path[arrowStyle] (c4)--node[rotate=-41, yshift=4mm]{founded\_in} node[rotate=-41, yshift=-4mm]{$\alpha_{i2}$} (us);
	
	\path[arrowStyle] (c1x)--node[rotate=15, yshift=4mm]{wife\_of} (c1);
	\path[arrowStyle] (c3x)--node[rotate=-15, yshift=4mm]{born\_in} (c3);
	
	\path[arrowStyle] (c1x)--node[rotate=-8, yshift=4mm]{\textbf{first\_lady}} node[rotate=-8, yshift=-3mm]{$\alpha_{i5}$} (us);
	\path[arrowStyle] (c3x)--node[rotate=8, yshift=4mm]{\textbf{native\_of}} node[rotate=8, yshift=-3mm]{$\alpha_{i6}$} (us);
	
	\node[entityId][fill=amber][right=11cm of us] (country) {$e_3$};
	\path[arrowStyle] (us)--node[yshift=2mm]{\LARGE $\rightarrow$} (country);
	
	\begin{scope}[every node/.style={entityId, fill=azure}]
		\node[above left=1cm of country, xshift=-5mm] (c1s) {$e_2$};
		\node[above right=1cm of country, xshift=5mm] (c2s) {$e_4$};
		\node[below left=1cm of country, xshift=-5mm] (c3s) {$e_6$};
		\node[below right=1cm of country, xshift=5mm] (c4s) {$e_7$};
		
		\node[below left=1.5cm of c1s, yshift=15mm, xshift=-5mm] (c1xs) {$e_1$};
		\node[above left=1.5cm of c3s, yshift=-15mm, xshift=-5mm] (c3xs) {$e_5$};
	\end{scope}
	\foreach \value in {1,...,4}
	\draw[->, line width=0.8mm, azure] (c\value s) -> (country);
	\foreach \value in {1,3}
	\draw[->, line width=0.8mm, azure] (c\value xs) -> (c\value s);
	
	\foreach \value in {1,3}
	\draw[->, line width=0.8mm, azure!60, dash pattern=on 3mm off 1mm, postaction={decorate}] (c\value xs) -> (country);
	
	\path[arrowStyle] (c1s)--node[rotate=-31, yshift=2mm]{$r_2$} (country);
	
	\path[arrowStyle] (c3s)--node[rotate=30, yshift=2mm]{$r_5$} (country);
	
	\path[arrowStyle] (c2s)--node[rotate=41, yshift=2mm]{$r_3$} (country);
	
	\path[arrowStyle] (c4s)--node[rotate=-41, yshift=2mm]{$r_6$} (country);
	
	\path[arrowStyle] (c1xs)--node[rotate=15, yshift=2mm]{$r_1$} (c1s);
	\path[arrowStyle] (c3xs)--node[rotate=-15, yshift=2mm]{$r_4$} (c3s);
	
	\path[arrowStyle] (c1xs)--node[rotate=-8, yshift=2mm]{$r_7$} (country);
	\path[arrowStyle] (c3xs)--node[rotate=8, yshift=2mm]{$r_8$} (country);
\end{tikzpicture}
	}
	\caption{Knowledge graph and normalized attention coefficients of the entity}
	\label{fig:graphExample}
\end{figure*}

\section{Related Work}
\label{chap:RelatedWork}

In this section, we present the basic definitions of knowledge graphs in order to understand the task of link prediction in knowledge graphs, as well as other related research directions.

\subsection{Definition of Knowledge Graphs}

The basic definitions of knowledge graphs are compiled and categorized by Cai, Hongyun \cite{cai2018comprehensive} and Goyal, Palash \cite{goyal2018graph} as follows:


\begin{itemize}
	\item \begin{definition}[Graph]\label{def:defGraph}
		\(\mathcal{G} = (V, E)\), where \(v \in V\) is a vertex and \(e \in E\) is an edge. \(\mathcal{G}\) is associated with a vertex-type mapping function \(f_v: V \to T^v\) and an edge-type mapping function \(f_e: E \to T^e\).
	\end{definition}
	
	Here, \(T^v\) and \(T^e\) are the sets of vertex types and edge types, respectively. Each vertex \(v_i \in V\) belongs to a specific type, i.e., \(f_v(v_i) \in T^v\). Similarly, for \(e_{ij} \in E, f_e (e_{ij}) \in T^e\).
	
	\item
	\begin{definition}[Homogeneous Graph]\label{def:homogeneous}
		Homogeneous graph: \textit{ $\mathcal{G}_{homo} = (V, E)$ is a graph where $\mid T^v \mid = \mid T^e \mid = 1$. All vertices in $\mathcal{G}$ belong to a single type, and all edges belong to a single type}.
	\end{definition}
	
	\item
	\begin{definition}[Heterogeneous Graph]\label{def:heterogeneous}
		Heterogeneous graph: \textit{$\mathcal{G}_{hete} = (V, E)$ is a graph where $\mid T^v \mid > 1$ or $\mid T^e \mid > 1$. That is, there is more than one type of vertex or more than one type of edge}.
	\end{definition}
	
	\item
	\begin{definition}[Knowledge Graph]\label{def:knowledgeGraph}
		Knowledge graph:
		$\mathcal{G}_{know} = (V, R, E)$ is a directed graph, where the vertex set represents entities, the relation set represents relationships between entities, and the edge set $E \subseteq V\times R \times V$ represents events in the form of subject-property-object triples. Each edge is a triple $(\text{entity}_{\text{head}}, \text{relation}, \text{entity}_{\text{tail}})$ (denoted as $\langle h, r, t \rangle$), expressing a relation $r$ from head entity $h$ to tail entity $t$.
	\end{definition}
	
	Here, $h, t \in V$ are entities and $r \in R$ is a relation. We refer to $\langle h, r, t \rangle$ as a knowledge graph triple.
	
	Example: in \autoref{fig:graphExample}, there are two triples: 
	$\langle \text{Tom Cruise, born\_in, New York} \rangle$
	and $\langle \text{New York, state\_of, U.S} \rangle$. Note that entities and relations in a knowledge graph often belong to different types. Therefore, a knowledge graph can be viewed as a specific case of a heterogeneous graph.
\end{itemize}

\subsection{Link Prediction in Knowledge Graphs}

Link prediction, also known as knowledge graph completion, is the task of exploiting known facts (events) in a knowledge graph to infer missing ones. This is equivalent to predicting the correct tail entity in a triple $\langle h, r, ? \rangle$ (tail prediction) or the correct head entity in $\langle ?, r, t \rangle$ (head prediction). For simplicity, instead of distinguishing between head and tail prediction, we generally refer to the known entity as the \textit{source entity} and the entity to be predicted as the \textit{target entity}.

Most current research on link prediction in knowledge graphs is related to approaches that focus on embedding a given graph into a low-dimensional vector space. In contrast to these approaches is a rule-based method explored in \cite{burl}. Its core algorithm samples arbitrary rules and generalizes them into Horn clauses \cite{wiki:Horn}, then uses statistics to compute the confidence of these generalized rules. When predicting a new link (edge) in the graph, the task is to infer whether an edge with a specific label exists between two given nodes. Many methods have been proposed to learn rules from graphs, such as in RuDiK \cite{ortona2018robust}, AMIE \cite{galarraga2015fast}, and RuleN \cite{meilicke2018fine}.

As mentioned earlier, there are two main approaches to this problem: one is optimizing an objective function to find a small set of rules that cover the majority of correct examples with minimal error, as explored in RuDiK \cite{ortona2018robust}. The other approach, which we adopt in this thesis, aims to explore all possible rules and then generate a top-\(k\) ranking of candidate triples, each associated with a confidence score measured on the training set.

In the deep learning branch of approaches, many successful techniques from image processing and natural language processing have been applied to knowledge graphs, such as Convolutional Neural Networks (CNNs \cite{lecun1999object}), Recurrent Neural Networks (RNNs \cite{hopfield2007hopfield}), and more recently, Transformers \cite{yang2019xlnet} and Capsule Neural Networks (CapsNets \cite{sabour2017dynamic}). In addition, other techniques such as random walks and hierarchical structure-based models have also been explored. The common advantage of these deep learning methods on knowledge graphs is their ability to automatically extract features and generalize complex graph structures based on large amounts of training data. However, some methods focus mainly on grid-like structures and fail to preserve the spatial characteristics of knowledge graphs.

The attention mechanism, particularly the multi-head attention layer, has been applied to graphs through the Graph Attention Network (GAT \cite{velivckovic2017graph}) model, which aggregates information about an entity based on attention weights from its neighboring entities. However, GAT lacks integration of relation embeddings and the embeddings of an entity's neighbors—components that are crucial for capturing the role of each entity. This limitation has been addressed in the work \textit{Learning Attention-based Embeddings for Relation Prediction in Knowledge Graphs} (\textbf{KBAT} \cite{nathani2019learning}), which we adopt as the foundation for our study.

The attention mechanism is currently one of the most effective (state-of-the-art) deep learning structures, as it has been proven to substitute any convolution operation \cite{cordonnier2019relationship}. Moreover, it serves as a core component in leading models for natural language processing, such as Megatron-LM \cite{shoeybi2019megatron}, and image segmentation, such as HRNet-OCR (Hierarchical Multi-Scale Attention \cite{tao2020hierarchical}). Some recent works \cite{cordonnier2020multi} have proposed interesting improvements based on the attention mechanism. However, these advancements have not yet been applied to knowledge graphs, which motivates us to adopt this family of methods to integrate the latest innovations into knowledge graph modeling.

\begin{figure*}[htbp]
	\centering
	
	\resizebox{1\textwidth}{!}{%
		\tikzset{
	category/.style  = {draw=none, thin, align=center},
	subcat/.style={rectangle, rounded corners=6pt},
	center/.style = {category, ellipse, fill=azure, text width=4em},
	group/.style = {category, subcat, fill=deepskyblue, rounded corners=3pt, text width=6em},
	yellowbox/.style = {category, subcat, fill=amber},
	greenbox/.style = {category, subcat, fill=darkpastelgreen},
	redbox/.style = {category, subcat, fill=awesome},
	bluebox/.style = {category, subcat, fill=deepmagenta!70},
	leafbox/.style = {category, subcat, fill=black!10, rounded corners=3pt}
}

\begin{tikzpicture}[>=Stealth]
	\node[center][execute at begin node=\setlength{\baselineskip}{0.9em}] (root) {Knowledge Graph};
	\node[group][above left=1cm of root, execute at begin node=\setlength{\baselineskip}{1em}] (c1) {Knowledge Representation Learning};
	\node[group][above right=8mm of root, execute at begin node=\setlength{\baselineskip}{1em}] (c2) {Knowledge Inference};
	\node[group][below left=5mm of root, execute at begin node=\setlength{\baselineskip}{1em}] (c3) {Knowledge Acquisition};
	\node[group][below right=20mm of root, xshift=-2cm, execute at begin node=\setlength{\baselineskip}{0.9em}] (c4) {Temporal Knowledge Graph};
	
	\begin{scope}[every node/.style={yellowbox}]
		\node[above=8mm of c1, xshift=-1cm] (c11) {Embedding Space};
		\node[above left=of c1, yshift=9mm, xshift=0mm, yshift=-18mm] (c12) {Scoring Function};
		\node[left=15mm of c1, xshift=5mm] (c13) {Model Encoding};
		\node[below left=of c1, xshift=10mm,yshift=0mm] (c14) {Auxiliary Information};
	\end{scope}
	
	\begin{scope}[every node/.style={redbox}]
		\node[above left=of c2, text width=35mm, yshift=1mm, xshift=14mm, execute at begin node=\setlength{\baselineskip}{0.9em}] (c21) {Natural Language Understanding};
		\node[above=1cm of c2, yshift=5mm, xshift=13mm] (c22) {Question Answering};
		\node[right=of c2, yshift=15mm] (c23) {Dialogue System};
		\node[right=of c2, yshift=4mm, xshift=7mm] (c24) {Recommendation System};
		\node[below=5mm of c2, yshift=-10mm, xshift=5mm] (c25) {Other Applications};
	\end{scope}
	
	\begin{scope}[every node/.style={greenbox}]
		\node[left=1cm of c3, yshift=0mm] (c31) {Entity Discovery};
		\node[below left=of c3, yshift=8mm, xshift=3mm] (c32) {Relation Extraction};
		\node[below right=10mm of c3, xshift=-4cm] (c33) {Graph Completion};
	\end{scope}
	
	\begin{scope}[every node/.style={bluebox}]
		\node[below=of c4, xshift=2mm, yshift=-15mm] (c41) {Temporal Logic Reasoning};
		\node[below=of c4, xshift=25mm, yshift=-5mm] (c42) {Time-Independent Relations};
		\node[below right=of c4, xshift=0mm, yshift=5mm] (c43) {Dynamic Entities};
		\node[right=of c4, xshift=0mm, yshift=-3mm] (c44) {Temporal Embeddings};
	\end{scope}
	
	\begin{scope}[every node/.style={leafbox}]
		\node[above=5mm of c11] (c11x) {
			\begin{tabular}{@{}l@{}@{}l@{}}
				- Point-wise & - Manifold \\
				- Complex & - Gaussian \\
				- Discrete & \\
			\end{tabular}
		};
		\node[above=5mm of c12, xshift=-10mm, yshift=6mm] (c12x) {
			\begin{tabular}{@{}l@{}}
				- Distance-based \\
				- Semantic \\
				- Others \\
			\end{tabular}
		};
		\node[left=of c13, yshift=5mm, xshift=3mm] (c13x) {
			\begin{tabular}{@{}l@{}}
				- Linear / Bilinear \\
				- Matrix Factorization \\
				- Neural Nets \\
				- CNN \\
				- RNN \\
				- Transformers \\
				- GCN \\
			\end{tabular}
		};
		\node[left=5mm of c14, xshift=0mm, yshift=-5mm] (c14x) {
			\begin{tabular}{@{}l@{}c@{}l@{}}
				- Textual & - Type & - Visual \\
			\end{tabular}
		};
		\node[below left=of c31, yshift=10mm] (c31x) {
			\begin{tabular}{@{}l@{}}
				- Recognition \\
				- Typing \\
				- Disambiguation \\
				- Ranking \\
			\end{tabular}
		};
		\node[below=of c32, xshift=-7mm] (c32x) {
			\begin{tabular}{@{}l@{}}
				- Neural Nets \\
				- Attention \\
				- GCN \\
				- GAN \\
				- RL \\
				- Others \\
			\end{tabular}
		};
		\node[below=5mm of c33, xshift=2mm] (c33x) {
			\begin{tabular}{@{}l@{}}
				- Ranking-based Embeddings \\
				- Textual Reasoning \\
				- Rule-based Reasoning \\
				- Hyper-relational Learning \\
				- Triple Classification \\
			\end{tabular}
		};
		\node[above=5mm of c22] (c22x) {
			\begin{tabular}{@{}l@{}}
				- Single-fact QA \\
				- Multi-step Reasoning \\
			\end{tabular}
		};
		\node[right=5mm of c25, yshift=3mm] (c25x) {
			\begin{tabular}{@{}l@{}}
				- Question Generation \\
				- Search Engines \\
				- Medical Applications \\
				- Health Recovery \\
				- Zero-shot Image Classification \\
				- Text Generation \\
				- Semantic Analysis \\
			\end{tabular}
		};
	\end{scope}
	
	\foreach \value in {1,...,4}
	\draw[->, very thick] (root) -> (c\value);
	
	\foreach \value in {1,...,4}
	\draw[->, thick] (c1) -> (c1\value);
	\foreach \value in {1,...,4}
	\draw[->, thick] (c1\value) -> (c1\value x);
	
	\foreach \value in {1,...,5}
	\draw[->, thick] (c2) -> (c2\value);
	\foreach \value in {2,5}
	\draw[->, thick] (c2\value) -> (c2\value x);
	
	\foreach \value in {1,...,3}
	\draw[->, thick] (c3) -> (c3\value);
	\foreach \value in {1,...,3}
	\draw[->, thick] (c3\value) -> (c3\value x);
	
	\foreach \value in {1,...,4}
	\draw[->, thick] (c4) -> (c4\value);
	
\end{tikzpicture}
	}
	\caption{
		A taxonomy of research areas in knowledge graphs}
	\label{fig:categoriesResearch}
\end{figure*}
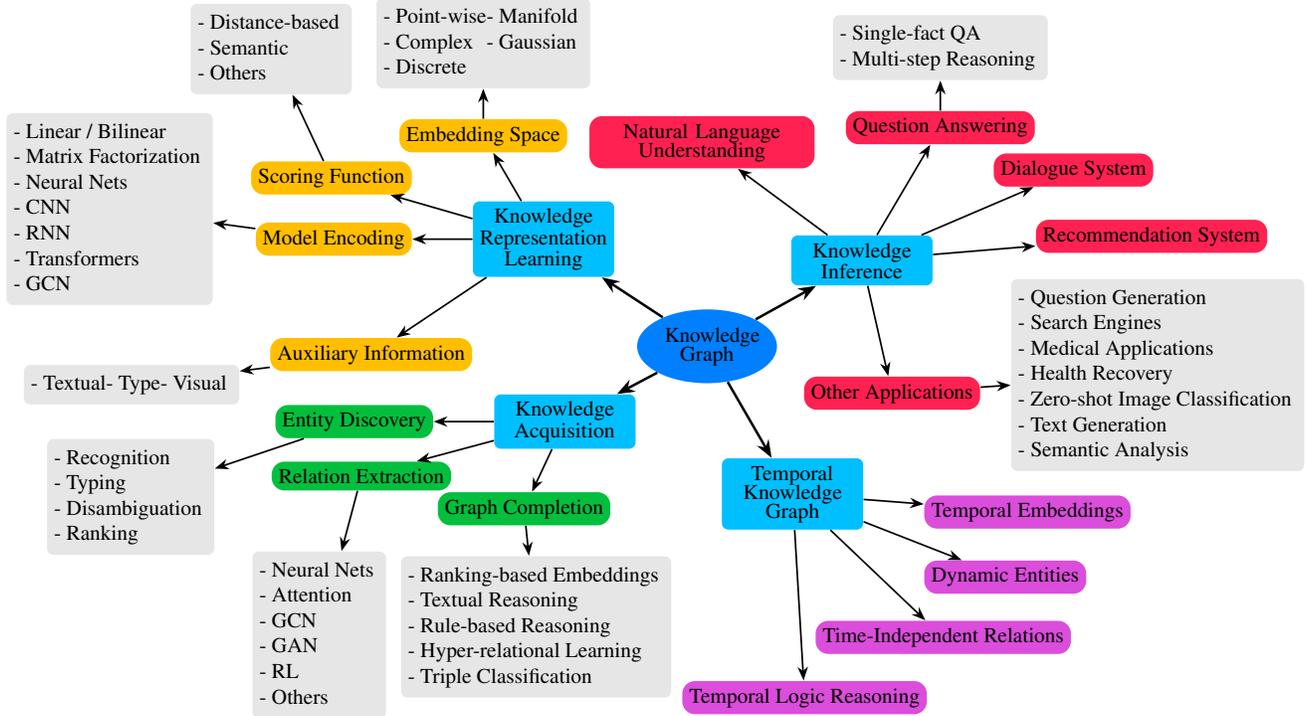

\subsection{Research Areas in Knowledge Graphs}

Knowledge representation has a long-standing history in logic and artificial intelligence. In the context of knowledge graphs, four major research areas have been categorized and summarized in the survey \cite{ji2020survey}, including: Knowledge Representation Learning, Knowledge Acquisition, Temporal Knowledge Graphs, and Knowledge-aware Applications. All research categories are illustrated in \autoref{fig:categoriesResearch}.

\textbf{Knowledge Representation Learning}

Knowledge representation learning is an essential research topic in knowledge graphs that enables a wide range of real-world applications. It is categorized into four subgroups:

\begin{itemize}
	\item \textit{Representation Space} focuses on how entities and relations are represented in vector space. This includes point-wise, manifold, complex vector space, Gaussian distribution, and discrete space embeddings.
	
	\item \textit{Scoring Function} studies how to measure the validity of a triple in practice. These scoring functions may be distance-based or similarity-based.
	
	\item \textit{Encoding Models} investigate how to represent and learn interactions among relations. This is currently the main research direction, including linear or non-linear models, matrix factorization, or neural network-based approaches.
	
	\item \textit{Auxiliary Information} explores how to incorporate additional information into embedding models, such as textual, visual, and type information.
\end{itemize}

\textbf{Knowledge Acquisition}

Knowledge acquisition focuses on how to extract or obtain knowledge based on knowledge graphs, including knowledge graph completion, relation extraction, and entity discovery. Relation extraction and entity discovery aim to extract new knowledge (relations or entities) into the graph from text. Knowledge graph completion refers to expanding an existing graph by inferring missing links. Research directions include embedding-based ranking, relation path reasoning, rule-based reasoning, and hyper-relational learning.

Entity discovery tasks include entity recognition, disambiguation, typing, and ranking. Relation extraction models often employ attention mechanisms, graph convolutional networks (GCNs), adversarial training, reinforcement learning (RL), deep learning, and transfer learning, which is the foundation of the method proposed in our work.

In addition, other major research directions in knowledge graphs include \textbf{temporal knowledge graphs} and \textbf{knowledge-aware applications}. Temporal knowledge graphs incorporate temporal information into the graph to learn temporal representations. Knowledge-aware applications include natural language understanding, question answering, recommendation systems, and many other real-world tasks where integrating knowledge improves representation learning.

\section{Deep Learning-Based Methods}
\label{chap:DeeLearning}

In this chapter, we present Knowledge Graphs and describe the task of Graph Embedding, providing an overview of current graph embedding techniques. We will revisit the attention mechanism and explain how it is applied to knowledge graphs through the Graph Attention Network (GAT) model \cite{velivckovic2017graph}. Additionally, we present an improved method based on the graph attention model - KBGAT \cite{nathani2019learning} - which incorporates relation information and neighboring relations.

\subsection{Graph Embedding}
\label{sec:graphEmbedding}

In the real world, representing entities and relations as vectors can be intuitively understood as the process of mapping features and attributes of an object into a lower-dimensional space, with each component representing a specific unit-level feature.

For example, we know Donald Trump is 1.9 meters tall and has a wife named Melania. Thus, we could represent the entity "Donald Trump" as a vector:

\[
\resizebox{\linewidth}{!}{$
\overrightarrow{e_\text{Trump}} = [1.9_{\text{height}}, 0_{\text{area}}, 1_{\text{wife is Melania}}, 0_{\text{wife is Taylor}}].
$}
\]

For features that cannot be measured or have no value (e.g., \texttt{.area}), we assign 0. For categorical features without magnitude (e.g., \texttt{.wife}), we represent them using probabilities of unit features (e.g., \texttt{.wife is Melania}, \texttt{.wife is Taylor}). Therefore, any object in the real world can be \textit{embedded} as a vector in an interpretable way.

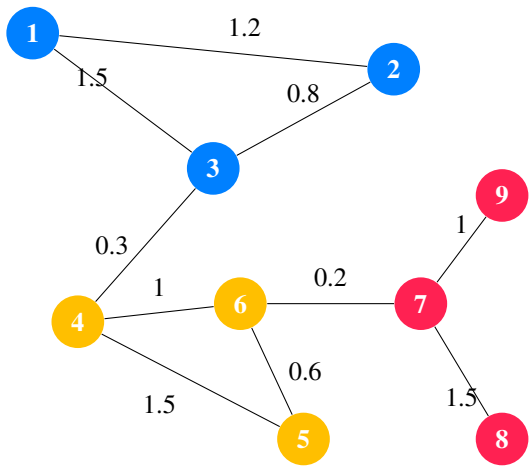
\begin{figure}[htp]
	\centering
	
	\begin{tikzpicture}[
	node/.style={circle, draw=none, color=white, minimum size=7mm,  font=\bfseries},
	green/.style={fill=azure},
	orange/.style={fill=amber},
	blue/.style={fill=awesome},
	every edge/.style={draw, thick},
	scale=1.2
	]
	
	\node[node, green] (1) at (0,5) {1};
	\node[node, green] (2) at (4,4.6) {2};
	\node[node, green] (3) at (2,3.5) {3};
	
	\node[node, orange] (4) at (0.5,1.8) {4};
	\node[node, orange] (5) at (3,0.5) {5};
	\node[node, orange] (6) at (2.3,2) {6};
	
	\node[node, blue] (7) at (4.3,2) {7};
	\node[node, blue] (8) at (5.2,0.5) {8};
	\node[node, blue] (9) at (5.2,3.2) {9};
	
	\draw (1) -- (3) node[midway, above left=1mm] {1.5};
	\draw (1) -- (2) node[midway, above right=1mm] {1.2};
	\draw (2) -- (3) node[midway, above=1mm] {0.8};
	
	\draw (3) -- (4) node[midway, left=1mm] {0.3};
	\draw (4) -- (6) node[midway, above=1mm] {1};
	\draw (4) -- (5) node[midway, below left=1mm] {1.5};
	\draw (5) -- (6) node[midway, right=1mm] {0.6};
	\draw (6) -- (7) node[midway, above=1mm] {0.2};
	\draw (7) -- (8) node[midway, below=1mm] {1.5};
	\draw (7) -- (9) node[midway, above=1mm] {1};
\end{tikzpicture}
	\caption{Example of an input graph}
	\label{fig:graphInput}
\end{figure}

To understand graph embedding techniques, we begin with several fundamental definitions:

\begin{itemize}
	\item
	\begin{definition}[First-Order Proximity]\label{def:firstOrderProximity}
		First-order proximity between vertex \(v_i\) and vertex \(v_j\) is the edge weight \(A_{i, j}\) of the edge \(e_{ij}\).
	\end{definition}
	
	Two vertices are more similar if they are connected by an edge with a higher weight. Thus, the first-order proximity between \(v_i\) and \(v_j\) is denoted as \(s^{(1)}_{ij} = A_{i, j}\). Let \(s^{(1)}_i = \begin{bmatrix} s^{(1)}_{i1}, s^{(1)}_{i2}, \dots, s^{(1)}_{i|V|} \end{bmatrix}\) represent the first-order proximities between \(v_i\) and other vertices.
	
	Using the graph in \autoref{fig:graphInput} as an example, the first-order proximity between \(v_1\) and \(v_2\) is the weight of edge \(e_{12}\), denoted as \(s^{(1)}_{12} = 1.2\). The vector \(s^{(1)}_1\) records the edge weights connecting \(v_1\) to all other vertices in the graph, i.e.,
	
	\[
	s^{(1)}_{1} = \begin{bmatrix} 0, 1.2, 1.5, 0, 0, 0, 0, 0, 0 \end{bmatrix}.
	\]
	
	\item
	\begin{definition}[Second-Order Proximity]\label{def:secondOrderProximity}
		Second-order proximity \(s^{(2)}_{ij}\) between vertex \(v_i\) and \(v_j\) is defined as the similarity between \(v_i'\)'s first-order neighborhood vector \(s^{(1)}_i\) and \(v_j'\)'s vector \(s^{(1)}_j\).
	\end{definition}
	
	For example, in \autoref{fig:graphInput}, the second-order proximity \(s^{(2)}_{12}\) is the similarity between \(s^{(1)}_1\) and \(s^{(1)}_2\). As introduced above:
	
	\[
	\resizebox{\linewidth}{!}{$
	s^{(1)}_1 = \begin{bmatrix} 0, 1.2, 1.5, 0, 0, 0, 0, 0, 0 \end{bmatrix}, \quad s^{(1)}_2 = \begin{bmatrix} 1.2, 0, 0.8, 0, 0, 0, 0 , 0, 0 \end{bmatrix}.
	$}
	\]
	
	We compute the cosine similarity:
	
	\[
	\resizebox{\linewidth}{!}{$
	s^{(2)}_{12} = \cos(s^{(1)}_1, s^{(1)}_2) = 0.43, \quad s^{(2)}_{15} = \cos(s^{(1)}_1, s^{(1)}_5) = 0.
	$}
	\]
	
	We observe that the second-order proximity between \(v_1\) and \(v_5\) is 0 because they share no common 1-hop neighbors. \(v_1\) and \(v_2\) share a common neighbor \(v_3\), thus their second-order proximity \(s^{(2)}_{12}\) is greater than 0.
	
	Higher-order proximities can be defined similarly. For example, the \(k\)-th order proximity between \(v_i\) and \(v_j\) is the similarity between \(s^{(k-1)}_i\) and \(s^{(k-1)}_j\).
	
	\item
	\begin{definition}[Graph Embedding]\label{def:graphEmbedding}
		Given a graph input \(\mathcal{G} = (V, E)\) and a predefined embedding dimension \(d\) where \(d \ll |V|\), the graph embedding problem is to map \(\mathcal{G}\) into a \(d\)-dimensional space while preserving as much graph property information as possible. These properties can be quantified using proximity measures such as first-order and higher-order proximity. Each graph is represented either as a \(d\)-dimensional vector (for the entire graph) or a set of \(d\)-dimensional vectors where each vector encodes a part of the graph (e.g., node, edge, substructure).
	\end{definition}
\end{itemize}

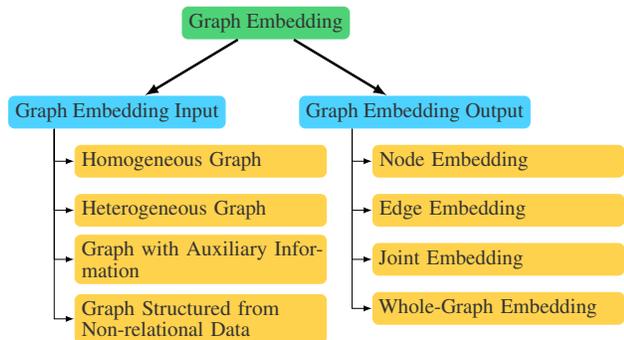
\begin{figure}[htp]
	\centering
	\normalsize
	\resizebox{\linewidth}{!}{%
		\begin{tikzpicture}[
	rec/.style  = {draw=none, rectangle, color=black!80,  execute at begin node=\setlength{\baselineskip}{1em}},
	root/.style = {rec, rounded corners=3pt,  fill=darkpastelgreen!70},
	level 1/.style={sibling distance=50mm},
	level 2/.style={rec, rounded corners=3pt, fill=deepskyblue!70},
	level 3/.style = {rec, rounded corners=3pt, align=left, fill=amber!70, text width=40mm, yshift=-10pt},
	edge from parent/.style={->,draw, very thick},
	>=latex]
	
	\node[root] {Graph Embedding}
	child {node[level 2] (c1) {Graph Embedding Input}}
	child {node[level 2] (c2) {Graph Embedding Output}};
	
	\begin{scope}[every node/.style={level 3}]
		\node [below of = c1, xshift=40pt, yshift=15pt] (c11) {Homogeneous Graph};
		\node [below of = c11, yshift=15pt] (c12) {Heterogeneous Graph};
		\node [below of = c12, yshift=15pt] (c13) {Graph with Auxiliary Information};
		\node [below of = c13, yshift=10pt] (c14) {Graph Structured from Non-relational Data};
		
		\node [below of = c2, xshift=40pt, yshift=15pt] (c21) {Node Embedding};
		\node [below of = c21, yshift=15pt] (c22) {Edge Embedding};
		\node [below of = c22, yshift=15pt] (c23) {Joint Embedding};
		\node [below of = c23, yshift=15pt] (c24) {Whole-Graph Embedding};
	\end{scope}
	
	\foreach \value in {1,...,4}
	\draw[->] (c1.195) |- (c1\value.west);
	
	\foreach \value in {1,...,4}
	\draw[->] (c2.195) |- (c2\value.west);
\end{tikzpicture}
	}
	\caption{Graph Embedding Techniques}
	\label{fig:graphEmbeddingSettingTree}
\end{figure}

Graph embedding is the process of transforming graph features into vectors or sets of low-dimensional vectors. The more effective the embedding, the higher the accuracy in subsequent graph mining and analysis tasks. The biggest challenge in graph embedding depends on the problem setting, which includes both the embedding input and output, as illustrated in \autoref{fig:graphEmbeddingSettingTree}.

Based on the embedding input, we categorize the surveyed methods in \cite{cai2018comprehensive} as follows: 
Homogeneous Graph, Heterogeneous Graph, Graph with Auxiliary Information, and Graph Constructed from Non-relational Data.

Different types of embedding inputs preserve different information in the embedding space and therefore pose different challenges for the graph embedding problem. 
For example, when embedding a graph with only structural information, the connections between nodes are the primary target to preserve. However, for graphs with node labels or entity attribute information, auxiliary information provides additional context for the graph and can therefore also be considered during the embedding process. Unlike embedding input, which is fixed and provided by the dataset, the embedding output is task-specific.

For instance, the most common embedding output is \textbf{node embedding}, which represents each node as a vector that reflects similarity between nodes. Node embeddings are beneficial for node-related tasks such as node classification, node clustering, etc.

However, in some cases, the tasks may involve more fine-grained graph components such as node pairs, subgraphs, or the entire graph. Therefore, the first challenge of embedding is to determine the appropriate type of embedding output for the application of interest. Four types of embedding outputs are illustrated in \autoref{fig:graphInput}, including: \textbf{Node Embedding} (\ref{fig:nodeEmbedding}), \textbf{Edge Embedding} (\ref{fig:edgeEmbedding}), \textbf{Hybrid Embedding} (\ref{fig:substructureEmbedding}), and \textbf{Whole-Graph Embedding} (\ref{fig:wholeGraphEmbedding}). Different output granularities have distinct criteria and present different challenges. For example, a good node embedding retains similarity with its neighbors in the embedding space. Conversely, a good whole-graph embedding represents the entire graph as a vector that preserves graph-level similarity.

\subsubsection{Graph Embedding Problem Settings}

While the input is determined by the type of information to be preserved, the output varies depending on the downstream graph mining task. Therefore, we discuss in more detail the embedding methods based on the type of output required by the embedding problem.

\begin{figure*}[htbp]
	\centering
	\begin{subfigure}[b]{0.48\textwidth}
		\centering
		\begin{tikzpicture}[scale=0.85]
	\begin{axis}[
		axis lines=box,
		xmin=0, xmax=3,
		ymin=-3, ymax=3,
		xtick={0, 0.75, 1.5, 2.25, 3},
		ytick={-3, -1.5, 0, 1.5, 3},
		grid=both,
		grid style={line width=0.3pt, draw=gray!30},
		major grid style={line width=0.4pt, draw=gray!50},
		ticklabel style={font=\small},
		width=\textwidth, height=\textwidth,
		enlargelimits=false,
		scatter/classes={
			a={mark=*,draw=black,fill=darkpastelgreen},
			b={mark=*,draw=black,fill=amber},
			c={mark=*,draw=black,fill=deepskyblue}
		},
		axis on top,
		clip=false
		]
		
		\addplot[
		scatter, only marks, scatter src=explicit symbolic, mark size=2.8pt
		] table[meta=class] {
			x     y     label  class
			0.4   1.8   1       a
			0.6   1.2   3       a
			0.8   2.2   2       a
			
			0.9   0.5   4       b
			1.0   0.0   5       b
			1.3   0.2   6       b
			
			1.7   0.3   7       c
			1.6  -0.3   8       c
			2.0  -0.2   9       c
		};
		
		\node[font=\bfseries] at (axis cs:0.4,2.1) {1};
		\node[font=\bfseries] at (axis cs:0.6,1.5) {3};
		\node[font=\bfseries] at (axis cs:0.8,2.5) {2};
		
		\node[font=\bfseries] at (axis cs:0.9,0.8) {4};
		\node[font=\bfseries] at (axis cs:1.0,0.3) {5};
		\node[font=\bfseries] at (axis cs:1.3,0.5) {6};
		
		\node[font=\bfseries] at (axis cs:1.7,0.6) {7};
		\node[font=\bfseries] at (axis cs:1.6,-0.0) {8};
		\node[font=\bfseries] at (axis cs:2.0,0.1) {9};
		
	\end{axis}
\end{tikzpicture}
		\caption{Node embedding with each vector representing node features}
		\label{fig:nodeEmbedding}
	\end{subfigure}
	\hfill
	\begin{subfigure}[b]{0.48\textwidth}
		\centering
		\begin{tikzpicture}[scale=0.85]
	\begin{axis}[
		axis lines=box,
		xmin=0, xmax=3,
		ymin=-3, ymax=3,
		xtick={0, 0.75, 1.5, 2.25, 3},
		ytick={-3, -1.5, 0, 1.5, 3},
		grid=both,
		grid style={line width=0.3pt, draw=gray!30},
		major grid style={line width=0.4pt, draw=gray!50},
		ticklabel style={font=\small},
		width=\textwidth, height=\textwidth,
		enlargelimits=false,
		scatter/classes={
			green={mark=triangle*,draw=black,fill=darkpastelgreen},
			orange={mark=triangle*,draw=black,fill=amber},
			blue={mark=triangle*,draw=black,fill=deepskyblue},
			yellow={mark=triangle*,draw=black,fill=amber},
			purple={mark=triangle*,draw=black,fill=deepmagenta}
		},
		axis on top,
		clip=false
		]
		
		\addplot[
		scatter, only marks, scatter src=explicit symbolic, mark size=4pt
		] table[meta=class] {
			x     y     label   class
			0.5   0.2   e_{13}  green
			0.7   1.1   e_{12}  green
			1.2   1.1   e_{23}  green
			
			0.8   0.3   e_{45}  orange
			1.0   0.0   e_{46}  orange
			1.8   0.1   e_{56}  orange
			
			0.4  -0.8   e_{78}  blue
			0.8  -1.3   e_{79}  blue
			
			2.0  -0.8   e_{67}  purple
			2.2  -1.0   e_{34}  yellow
		};
		
		\node[font=\bfseries] at (axis cs:0.5,0.45) {e\textsubscript{13}};
		\node[font=\bfseries] at (axis cs:0.7,1.35) {e\textsubscript{12}};
		\node[font=\bfseries] at (axis cs:1.2,1.35) {e\textsubscript{23}};
		
		\node[font=\bfseries] at (axis cs:0.8,0.55) {e\textsubscript{45}};
		\node[font=\bfseries] at (axis cs:1.0,0.3) {e\textsubscript{46}};
		\node[font=\bfseries] at (axis cs:1.8,0.35) {e\textsubscript{56}};
		
		\node[font=\bfseries] at (axis cs:0.4,-0.55) {e\textsubscript{78}};
		\node[font=\bfseries] at (axis cs:0.8,-1.05) {e\textsubscript{79}};
		
		\node[font=\bfseries] at (axis cs:2.0,-0.55) {e\textsubscript{67}};
		\node[font=\bfseries] at (axis cs:2.2,-0.75) {e\textsubscript{34}};
		
	\end{axis}
\end{tikzpicture}
		\caption{Edge embedding with each vector representing edge features}
		\label{fig:edgeEmbedding}
	\end{subfigure}
	
	\vspace{0.5em}
	
	\begin{subfigure}[b]{0.48\textwidth}
		\centering
		\begin{tikzpicture}[scale=0.85]
    \begin{axis}[
        axis lines=box,
        xmin=0, xmax=3,
        ymin=-3, ymax=3,
        xtick={0, 0.75, 1.5, 2.25, 3},
        ytick={-3, -1.5, 0, 1.5, 3},
        grid=both,
        grid style={line width=0.3pt, draw=gray!30},
        major grid style={line width=0.4pt, draw=gray!50},
        ticklabel style={font=\small},
        width=\textwidth, height=\textwidth,
        enlargelimits=false,
        scatter/classes={
            green={mark=diamond*,draw=black,fill=darkpastelgreen},
            orange={mark=diamond*,draw=black,fill=amber},
            blue={mark=diamond*,draw=black,fill=deepskyblue}
        },
        axis on top,
        clip=false
        ]
        
        \addplot[
        scatter, only marks, scatter src=explicit symbolic, mark size=4pt
        ] table[meta=class] {
            x     y     label       class
            2.0   0.5   G_{1,2,3}    green
            1.5   0.0   G_{4,5,6}    orange
            1.0  -1.5   G_{7,8,9}    blue
        };
        
        \node[font=\bfseries\Large] at (axis cs:2.0,0.85)  {G\textsubscript{\{1,2,3\}}};
        \node[font=\bfseries\Large] at (axis cs:1.5,0.35)  {G\textsubscript{\{4,5,6\}}};
        \node[font=\bfseries\Large] at (axis cs:1.0,-1.15) {G\textsubscript{\{7,8,9\}}};
        
    \end{axis}
\end{tikzpicture}
		\caption{Embedding a graph substructure}
		\label{fig:substructureEmbedding}
	\end{subfigure}
	\hfill
	\begin{subfigure}[b]{0.48\textwidth}
		\centering
		\begin{tikzpicture}[scale=0.85]
        \begin{axis}[
            axis lines=box,
            xmin=0, xmax=3,
            ymin=-3, ymax=3,
            xtick={0, 0.75, 1.5, 2.25, 3},
            ytick={-3, -1.5, 0, 1.5, 3},
            grid=both,
            grid style={line width=0.3pt, draw=gray!30},
            major grid style={line width=0.4pt, draw=gray!50},
            ticklabel style={font=\small},
            width=\textwidth, height=\textwidth,
            enlargelimits=false,
            axis on top,
            clip=false
            ]
            
            \node[regular polygon, regular polygon sides=6, minimum size=8pt,
            fill=black, draw=black] at (axis cs:1.2,0.4) {};
            
            \node[font=\bfseries\Large] at (axis cs:1.0,0.0) {G\textsubscript{1}};
            
        \end{axis}
    \end{tikzpicture}
		\caption{Whole-graph embedding}
		\label{fig:wholeGraphEmbedding}
	\end{subfigure}
	\caption{Graph embedding methods.}
	\label{fig:GraphEmbeddingMethods}
\end{figure*}
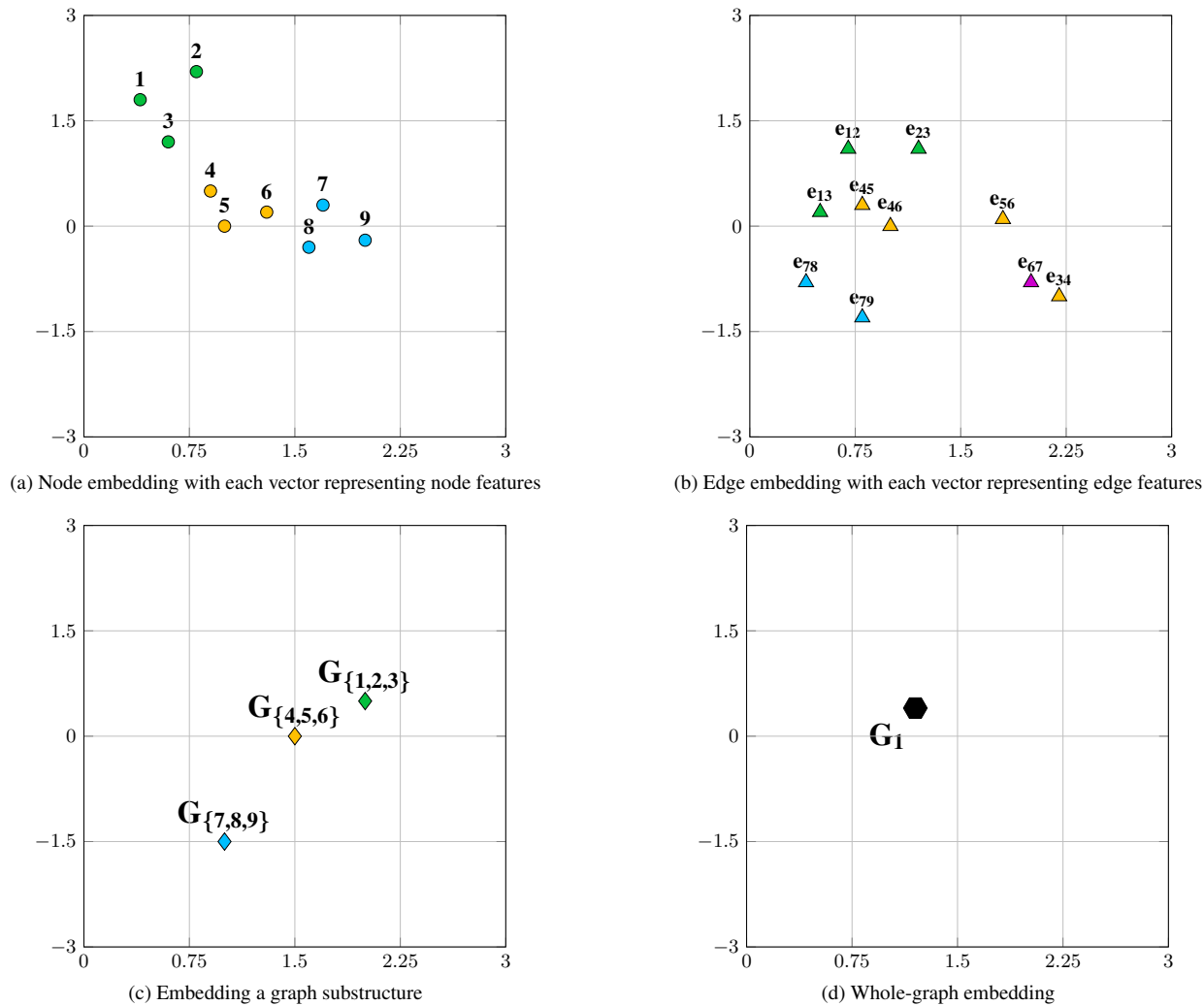

%
%
%
%
%
%
%
%
%

\textbf{Node Embedding}
\label{sec:nodeEmbedding}

Node embedding \autoref{fig:nodeEmbedding} represents each node as a low-dimensional vector. Nodes that are \textit{close} in the graph have similar vector representations. The difference among graph embedding methods lies in how they define the \textit{closeness} between two nodes. First-order proximity (Definition \ref{def:firstOrderProximity}) and second-order proximity (Definition \ref{def:secondOrderProximity}) are two commonly used metrics to measure pairwise node similarity. Higher-order proximity has also been explored to some extent. For example, capturing k-step (k = 1, 2, 3, ···) neighborhood relationships during embedding is discussed in the study by Cao, Shaosheng \cite{cao2015grarep}.

\textbf{Edge Embedding}
\label{sec:edgeEmbedding}

In contrast to node embedding, edge embedding \autoref{fig:edgeEmbedding} aims to represent an edge as a low-dimensional vector. Edge embedding is useful in two cases:

First, in knowledge graph embedding. Each edge is a triple $\langle h, r, t \rangle$ (Definition \ref{def:knowledgeGraph}). The embedding is learned to preserve the relation $r$ between $h$ and $t$ in the embedding space so that a missing entity or relation can be accurately predicted given the other two components in $\langle h, r, t \rangle$.

Second, some methods embed a pair of nodes as a vector to compare it with other node pairs or to predict the existence of a link between the two nodes. Edge embedding benefits graph analyses that involve edges (node pairs), such as link prediction, relation prediction, and knowledge-based entity inference.

\textbf{Hybrid Embedding}

Hybrid embedding refers to embedding combinations of different graph components, e.g., node + edge (i.e., substructure), or node + parts. Substructure or part embeddings \autoref{fig:substructureEmbedding} can also be derived by aggregating individual node and edge embeddings. However, such \textit{indirect} approaches are not optimized to capture the structure of the graph. Moreover, node and part embeddings can reinforce each other. Node embeddings improve by learning from high-order neighborhood attention, while part embeddings become more accurate due to the collective behavior of their constituent nodes.

\textbf{Whole-Graph Embedding}

Whole-graph embedding \autoref{fig:wholeGraphEmbedding} is typically used for small graphs such as proteins, molecules, etc. In this case, an entire graph is represented as a vector, and similar graphs are embedded close to each other. Whole-graph embedding is useful for graph classification tasks by providing a simple and effective solution for computing graph similarity. To balance embedding time (efficiency) and information retention (expressiveness), \textbf{hierarchical graph embedding} \cite{mousavi2017hierarchical} introduces a hierarchical embedding framework. It argues that accurate understanding of global graph information requires processing substructures at multiple scales. A graph pyramid is formed where each level is a coarsened graph at a different scale. The graph is embedded at all levels and then concatenated into a single vector. Whole-graph embedding requires collecting features from the entire graph, thus is generally more time-consuming than other settings.

\subsubsection{Graph Embedding Techniques}

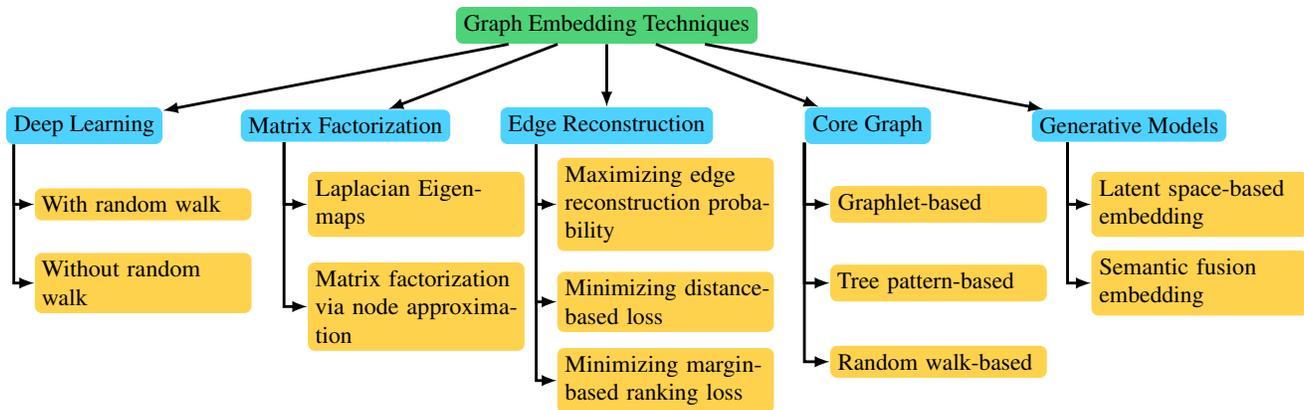
\begin{figure*}[htp]
	\centering
		\resizebox{1\textwidth}{!}{%
			\begin{tikzpicture}[
	rec/.style  = {draw=none, rectangle, thin, execute at begin node=\setlength{\baselineskip}{1.2em}},
	root/.style = {rec, rounded corners=3pt, align=center, fill=darkpastelgreen!70},
	level 1/.style={sibling distance=39mm},
	level 2/.style={rec, rounded corners=3pt, fill=deepskyblue!70},
	level 3/.style = {rec, align=left, fill=amber!70, text width=30mm, yshift=-5pt, rounded corners=3pt},
	edge from parent/.style={->,draw, very thick},
	>=latex]
	
	\node[root] {Graph Embedding Techniques}
	child {node[level 2] (c1) {Deep Learning}}
	child {node[level 2] (c2) {Matrix Factorization}}
	child {node[level 2] (c3) {Edge Reconstruction}}
	child {node[level 2] (c4) {Core Graph}}
	child {node[level 2] (c5) {Generative Models}};
	
	\begin{scope}[every node/.style={level 3}]
		\node [below of = c1, xshift=25pt] (c11) {With random walk};
		\node [below of = c11] (c12) {Without random walk};
		
		\node [below of = c2, xshift=30pt] (c21) {Laplacian Eigenmaps};
		\node [below of = c21, yshift=-10pt] (c22) {Matrix factorization via node approximation};
		
		\node [below of = c3, xshift=25pt] (c31) {Maximizing edge reconstruction probability};
		\node [below of = c31, yshift=-8pt] (c32) {Minimizing distance-based loss};
		\node [below of = c32] (c33) {Minimizing margin-based ranking loss};
		
		\node [below of = c4, xshift=30pt] (c41) {Graphlet-based};
		\node [below of = c41] (c42) {Tree pattern-based};
		\node [below of = c42] (c43) {Random walk-based};
		
		\node [below of = c5, xshift=30pt] (c51) {Latent space-based embedding};
		\node [below of = c51] (c52) {Semantic fusion embedding};
	\end{scope}
	
	\foreach \value in {1,2}
	\draw[->, very thick] (c1.195) |- (c1\value.west);
	
	\foreach \value in {1, 2}
	\draw[->, very thick] (c2.195) |- (c2\value.west);
	
	\foreach \value in {1,...,3}
	\draw[->, very thick] (c3.195) |- (c3\value.west);
	
	\foreach \value in {1,...,3}
	\draw[->, very thick] (c4.195) |- (c4\value.west);
	
	\foreach \value in {1,...,2}
	\draw[->, very thick] (c5.195) |- (c5\value.west);
\end{tikzpicture}
	}
	\caption{Graph Embedding Techniques}
	\label{fig:graphEmbeddingTechniquesTree}
\end{figure*}

In this section, we categorize graph embedding methods based on the techniques used. As previously stated, the goal of graph embedding is to represent a graph in a lower-dimensional space while preserving as much of the original graph information as possible. The main differences among embedding techniques lie in how they define which intrinsic graph properties should be preserved. Since the primary focus of our work is on graph embedding methods based on deep learning, we provide only brief overviews of the other method categories.

\textbf{Deep Learning}

In this section, we present in detail the research directions of deep learning techniques, including: using random walks and not using random walks. Deep learning techniques are widely used in graph embedding due to their speed and efficiency in automatically capturing features. Among these deep learning-based methods, three types of input-based graph settings (excluding graphs constructed from non-relational data) and all four output types (as shown in \autoref{fig:graphEmbeddingSettingTree}) can adopt deep learning approaches.

\textit{Deep learning techniques with random walks}

In this category, the second-order proximity (Definition \ref{def:secondOrderProximity}) in the graph is preserved in the embedding space by maximizing the probability of observing a vertex's neighborhood conditioned on its embedding vector. The graph is represented as a set of samples obtained via random walks, and then deep learning methods are applied to ensure the structural properties (i.e., path-based information) are preserved. Representative methods in this group include: DeepWalk \cite{perozzi2014deepwalk}, LINE \cite{tang2015line}, Node2Vec \cite{grover2016node2vec}, Anonymous Walk \cite{ivanov2018anonymous}, NetGAN \cite{bojchevski2018netgan}, etc.

\textit{Deep learning techniques without random walks}

This approach applies multi-layer learning structures effectively and efficiently to transform the graph into a lower-dimensional space. It operates over the entire graph. Several popular methods have been surveyed and presented in \cite{rossi2020knowledge}, as follows:

\begin{itemize}
	\item \textbf{Convolutional Neural Networks (CNNs)}
	
	This model uses multiple convolutional layers: each layer performs convolution on the input data using low-dimensional filters. The result is a feature map, which then passes through a fully connected layer to compute the probability values. For example, \textbf{ConvE} \cite{dettmers2017convolutional}: each entity and relation is represented as a low-dimensional vector ($d$-dimensional). For each triple, it concatenates and reshapes the head $h$ and relation $r$ embeddings into a single input $[h, r]$ with shape $d_m \times d_n$. This is then passed through a convolutional layer with a filter $\omega$ of size $m \times n$, followed by a fully connected layer with weights $W$. The final result is combined with the tail embedding $t$ using a dot product. This architecture can be considered a \textit{multi-class classification} model.
	
	Another popular model is \textbf{ConvKB} \cite{nguyen2017novel}, which is similar to ConvE but concatenates the three embeddings $h$, $r$, and $t$ into a matrix $[h, r, t]$ of shape $d \times 3$. It is then passed through a convolutional layer with $T$ filters of size $1 \times 3$, resulting in a feature map of size $T \times 3$. This is further passed through a fully connected layer with weights $\mathbf{W}$. This architecture can be considered a binary classification model.
	
	\item \textbf{Recurrent Neural Networks (RNNs)}
	
	These models apply one or more recurrent layers to analyze the entire path (a sequence of events/triples) sampled from the training set, instead of treating each event independently. For example, RSN \cite{guo2019learning} notes that traditional RNNs are unsuitable for graphs because each step only takes the relation information without considering the entity embedding from the previous step. Therefore, it fails to clearly model the transitions among entity-relation paths. To address this, they propose RSN (Recurrent Skipping Networks \cite{guo2019learning}): at each step, if the input is a relation, a hidden state is updated to reuse the entity embedding. The output is then dot-multiplied with the target embedding vector.
	
	\item \textbf{Capsule Neural Networks}
	
	Capsule networks group neurons into "capsules" \label{capsule}, where each capsule encodes specific features of the input, such as representing a particular part of an image. One advantage of capsule networks is their ability to capture spatial relationships that are lost in conventional convolution. Each capsule produces feature vectors. For instance, \textbf{CapsE} \cite{vu2019capsule}: each entity and relation is embedded into vectors, similar to ConvKB. It concatenates the embeddings $h$, $r$, and $t$ into a matrix of shape $d \times 3$, then applies $E$ convolutional filters of size $1 \times 3$, resulting in a $d \times E$ matrix. Each $i$-th row encodes distinct features of $h[i]$, $r[i]$, and $t[i]$. This matrix is then fed into a capsule layer, where each capsule (\ref{capsule}) processes a column, thus receiving feature-specific information from the input triple. A second capsule layer is used to produce the final output.
	
	\item \textbf{Graph Attention Networks (GATs)}
	
	This category uses the attention mechanism \cite{vaswani2017attention}, which has achieved notable success in NLP tasks. For each embedding vector, information from neighboring entities is aggregated using attention weights. These are then combined and passed through a fully connected layer with learnable weights to obtain the final embeddings. For example, GAT \cite{velivckovic2017graph} applies multi-head attention to each training triple to generate an embedding vector. This embedding is then transformed via a weight matrix to produce a higher-dimensional vector that aggregates information from neighboring nodes in the original triple. An improved version, KBGAT \cite{nathani2019learning}, incorporates the relation embedding into the attention mechanism. These methods will be discussed in detail in the subsequent sections.
	
	\item \textbf{Other methods}
	
	There are also other approaches, such as autoencoder-based techniques like Structural Deep Network Embedding (SDNE) \cite{wang2016structural}.
\end{itemize}

\textbf{Matrix Factorization}

Matrix factorization-based graph embedding represents the structural characteristics of a graph (e.g., similarity or proximity between vertex pairs) in the form of a matrix and then factorizes this matrix to obtain vertex embeddings. The input for this category of methods is typically high-dimensional non-relational features, and the output is a set of vertex embeddings. There are two matrix factorization-based graph embedding methods: Graph Laplacian Eigenmaps and Node Proximity Matrix Factorization.

\begin{itemize}
	\item \textit{Graph Laplacian Eigenmaps}
	
	This approach preserves graph properties by analyzing similar vertex pairs and heavily penalizes embeddings that place highly similar nodes far apart in the embedding space.
	
	\item \textit{Node Proximity Matrix Factorization}
	
	This approach approximates neighboring nodes in a low-dimensional space using matrix factorization techniques. The objective is to preserve neighborhood information by minimizing the approximation loss.
\end{itemize}

\textbf{Edge Reconstruction}

The edge reconstruction method builds edges based on the vertex embeddings so that the reconstructed graph is as similar as possible to the input graph. This method either maximizes the edge reconstruction probability or minimizes edge reconstruction loss. Additionally, the loss can be distance-based or margin-based ranking loss.

\begin{itemize}
	\item \textit{Maximize Edge Reconstruction Probability}
	
	In this method, a good vertex embedding maximizes the likelihood of generating observed edges in the graph. In other words, a good vertex embedding should allow for reconstructing the original input graph. This is achieved by maximizing the generative probability of all observed edges using vertex embeddings.
	
	\item \textit{Minimize Distance-Based Loss}
	
	In this approach, embeddings of neighboring nodes should be as close as possible to the observed neighboring nodes in the original graph. Specifically, vertex proximity can be measured using their embeddings or heuristically based on observed edges. The difference between these two types of proximity is then minimized to ensure consistent similarity.
	
	\item \textit{Minimize Margin-Based Ranking Loss}
	
	In this approach, the edges in the input graph represent correlations between vertex pairs. Some vertices in the graph are often linked with related vertex sets. This method ensures that embeddings of related nodes are closer together than unrelated ones by minimizing a margin-based ranking loss.
\end{itemize}

\textbf{Graph Kernels}

Graph kernel methods represent the entire graph structure as a vector containing counts of basic substructures extracted from the graph. Subcategories of graph kernel techniques include: graphlets, subtree patterns, and random walk-based methods.

This approach is designed to embed whole graphs, focusing only on global graph features. The input is typically homogeneous graphs or graphs with auxiliary information.

\textbf{Generative Models}

A generative model is defined by specifying a joint distribution over input features and class labels, parameterized by a set of variables. There are two subcategories of generative model-based methods: embedding graphs into latent space and incorporating semantics for embedding. Generative models can be applied to both node and edge embeddings. They are commonly used to embed semantic information, with inputs often being heterogeneous graphs or graphs with auxiliary attributes.

\begin{itemize}
	\item \textit{Embedding Graphs into Latent Semantic Space}
	
	In this group, vertices are embedded into a latent semantic space where the distance between nodes captures the graph structure.
	
	\item \textit{Incorporating Semantics for Embedding}
	
	In this method, each vertex is associated with graph semantics and should be embedded closer to semantically relevant vertices. These semantic relationships can be derived from descriptive nodes via a generative model.
\end{itemize}

\textbf{Summary}: Each graph embedding method has its own strengths and weaknesses, which have been summarized by Cai, Hongyun \cite{cai2018comprehensive}. The \textit{matrix factorization} group learns embeddings by analyzing pairwise global similarities. The \textit{deep learning} group, in contrast, achieves promising results and is suitable for graph embedding because it can learn complex representations from complex graph structures.

Graph embedding methods based on random walk in deep learning have lower computational cost compared to those using full deep learning models. Traditional methods often treat graphs as grids; however, this does not reflect the true nature of graphs. In the \textit{edge reconstruction} group, the objective function is optimized based on observed edges or by ranking triplets. While this approach is more efficient, the resulting embedding vectors do not account for the global structure of the graph. The \textit{graph kernel} methods transform graphs into vectors, enabling graph-level tasks such as graph classification. Therefore, they are only effective when the desired primitive structures in a graph can be enumerated. The \textit{generative model} group naturally integrates information from multiple sources into a unified model. Embedding a graph into a latent semantic space produces interpretable embedding vectors using semantics. However, assuming the modeling of observations using specific distributions can be difficult to justify. Moreover, generative approaches require a large amount of training data to estimate a model that fits the data well. Hence, they may perform poorly on small graphs or when only a few graphs are available.

Among these methods, deep learning-based graph embedding allows learning complex representations and has shown the most promising results. Graph attention networks (GATs), which are based on attention mechanisms, aggregate the information of an entity using attention weights from its neighbors relative to the central entity. We believe this research direction is aligned with studies on the relationship between attention and memory \cite{memoryandattention:2020}, where the distribution of attention determines the weight or importance of one entity relative to another. Likewise, the embedding vector representing an entity is influenced by the attention or importance of its neighboring embeddings. Therefore, this is the approach we selected among the existing graph embedding methods.

\subsection{Multi-head Attention Mechanism}

In 2014, the multi-head attention mechanism was introduced by Bahdanau, Dzmitry \cite{bahdanau2014neural}, but it only gained widespread popularity in 2017 through the Transformer model by Vaswani, Ashish \cite{vaswani2017attention}. The attention mechanism is an effective method to indicate the importance of a word with respect to other words in a sentence, and it has been shown to be a generalization of any convolution operation as reported by Cordonnier, Jean-Baptiste \cite{cordonnier2019relationship}. To understand how multi-head attention is applied to graphs, in this section we will detail the mechanism so we can better understand how it is used in link prediction tasks within knowledge graphs.

\subsubsection{Attention Mechanism}
\label{sec:attentionMechanism}

The input of the attention mechanism consists of two embedding matrices $\mathbf{X} = \Big\{\overrightarrow{x_1}, \overrightarrow{x_2}, ...,  \overrightarrow{x_{N_x}}\Big\}$ and $\mathbf{Y} = \Big\{\overrightarrow{y_1}, \overrightarrow{y_2}, ...,  \overrightarrow{y_{N_y}}\Big\}$, where each row $i^{\text{th}}$ or $j^{\text{th}}$ in matrix $\mathbf{X}$ or $\mathbf{Y}$ is an embedding vector $\overrightarrow{x_i} \in \mathbb{R}^{1 \times D_{\text{in}}}$, $\overrightarrow{y_j} \in \mathbb{R}^{1 \times D_{\text{in}}}$.

The attention mechanism transforms input vectors of $D_{\text{in}}$ dimensions into output vectors of $D_{\text{attention}}$ dimensions to represent the importance of each of the $N_x$ elements $x$ with respect to all $N_y$ elements $y$. Given $\mathbf{X} \in \mathbb{R}^{N_x \times D_\text{in}}$ and $\mathbf{Y} \in \mathbb{R}^{N_y \times D_\text{in}}$ as input embedding matrices, and $\mathbf{H} \in \mathbb{R}^{N_x \times D_\text{attention}}$ as the output embedding matrix, the attention mechanism introduced by Vaswani et al. \cite{vaswani2017attention} is defined as follows:

\begin{equation}
	\label{attention}
	\mathbf{H} = \text{Attention}(\mathbf{Q}, \mathbf{K}, \mathbf{V}) = \text{softmax}\Big(\frac{\mathbf{Q}\mathbf{K}^T}{\sqrt{d_k}}\Big) \mathbf{V}
\end{equation}

where $\mathbf{Q} = \mathbf{X}\mathbf{W}_Q, \mathbf{K} = \mathbf{Y} \mathbf{W}_K, \mathbf{V} = \mathbf{Y} \mathbf{W}_V$.

The weight matrices 
$\mathbf{W}_Q \in \mathbb{R}^{D_{\text{in}} \times D_{k}}$, 
$\mathbf{W}_K \in \mathbb{R}^{D_{\text{in}} \times D_{k}}$, and 
$\mathbf{W}_V \in \mathbb{R}^{D_{\text{in}} \times D_{\text{attention}}}$ 
are used to parameterize the transformation from input embedding vectors of dimension $D_{\text{in}}$ into output embedding vectors of dimension $D_k$ or $D_{\text{attention}}$. The term $\mathbf{Q}\mathbf{K}^T$ represents the dot product between each vector $x$ and all vectors $y$. Dividing by $\sqrt{d_k}$ normalizes the result with respect to the vector dimension $k$. The result is then passed through the \textit{softmax} function to enable comparison of attention scores across different pairs. We can interpret $\text{softmax}\Big(\frac{\mathbf{Q}\mathbf{K}^T}{\sqrt{d_k}}\Big)$ as the \textit{attention coefficients}, indicating the importance of each $y$ with respect to each $x$. Finally, this is multiplied with the value matrix $\mathbf{V}$ to produce the final output embedding of dimension $D_{\text{attention}}$.

If $\mathbf{X} = \mathbf{Y}$, we are computing the importance of each element with respect to other elements in the same input matrix, which is referred to as the self-attention mechanism.

\subsubsection{Multi-Head Attention}

The multi-head attention mechanism is a way of combining multiple attention layers to stabilize the learning process. Similar to the standard attention mechanism above, the multi-head attention mechanism transforms the initial $N_x$ embedding vectors of dimension $D_{\text{in}}$ into output embedding vectors of dimension $D_{\text{multi-head}}$, aggregating information from various other nodes to provide greater stability during training. Multi-head attention stacks $N_{\text{head}}$ attention output matrices $\mathbf{H}$, and then applies a weight matrix to transform the original embedding matrix $\mathbf{X} \in \mathbb{R}^{N_x \times D_\text{in}}$ into a new embedding matrix $\mathbf{X}' \in \mathbb{R}^{N_x \times D_{\text{multi-head}}}$ using the following formula:

\begin{equation}
	\label{headAttention}
	\begin{split}
		\mathbf{X}'& =\left(\bigparallel_{h=1}^{N_{\text{head}}}\mathbf{H}^{(h)}\right)\mathbf{W}^{O} \\
		& = \left(\bigparallel_{h=1}^{N_{\text{head}}} \text{Attention}(\mathbf{X} \mathbf{W}_Q^{(h)}, \mathbf{Y} \mathbf{W}_K^{(h)}, \mathbf{Y} \mathbf{W}_V^{(h)}) \right)\mathbf{W}^{O}
	\end{split}
\end{equation}

Here, the weight matrices $\mathbf{W}_Q^{(h)}$, $\mathbf{W}_K^{(h)} \in \mathbb{R}^{D_{\text{in}} \times D_{k}}$, and $\mathbf{W}_V^{(h)} \in \mathbb{R}^{D_{\text{in}} \times D_{\text{attention}}}$ correspond to each individual attention head $h \in [N_{\text{head}}]$. The output projection matrix $\mathbf{W}^{O} \in \mathbb{R}^{N_{\text{head}} D_{\text{attention}} \times D_{\text{multi-head}}}$ parameterizes the transformation of the concatenated output heads into the final output embedding matrix.

At this point, we have presented the attention mechanism for computing attention scores and aggregating embedding information from neighboring vectors. In the next section, we will describe how this attention mechanism is applied to knowledge graphs.



\subsection{Graph Attention Network}
\label{sec:GAT}

With the success of the \textit{multi-head attention mechanism} in natural language processing, it has also been studied for applications in image processing models \cite{ramachandran2019stand}. Consequently, the multi-head attention mechanism has been explored for replacing convolutional operations in knowledge graph embedding models, such as Graph Convolutional Networks (GCNs \cite{kipf2016semi}). In this section, we present in detail how the attention mechanism from \ref{sec:attentionMechanism} is applied to graph embedding via the Graph Attention Network (GAT \cite{velivckovic2017graph}) method.

The input to the \textit{graph attention network} model is a set of embedding vectors, randomly initialized from a normal distribution, representing features of each entity: $\mathbf{E} = \Big\{\overrightarrow{e_1}, \overrightarrow{e_2}, ...,  \overrightarrow{e_{N_e}}\Big\}$. The objective of the model is to transform this into a new output embedding matrix $\mathbf{E}'' = \Big\{\overrightarrow{e''_1}, \overrightarrow{e''_2}, ...,  \overrightarrow{e''_{N_e}}\Big\}$ capable of aggregating embedding information from neighboring entities. Here, $\mathbf{E} \in \mathbb{R}^{N_e \times D_{\text{in}}}$ and $\mathbf{E}'' \in \mathbb{R}^{N_e \times D''}$ denote the input and output embedding matrices for the entity set, respectively. $N_e$ is the number of entities, and $D_{\text{in}}$, $D''$ are the dimensions of the input and output embeddings.

Similar to the multi-head attention mechanism introduced in \ref{sec:attentionMechanism}, the application of this mechanism to a knowledge graph follows the same logic as the \textit{self-attention mechanism}, in which each node attends to all other nodes in the graph. However, computing attention scores between every pair of nodes in a graph is not meaningful if no relationship exists between them, and it would incur significant computational overhead. Therefore, the model applies a technique known as \textit{masked attention}, in which all attention scores corresponding to unrelated nodes in the graph are ignored. These relevant connections are precisely defined as the first-order proximity (Definition \ref{def:firstOrderProximity}) of a node in the graph. Thus, in this context, we let $\mathbf{X} = \mathbf{Y} = \mathbf{E}$ (as in \ref{sec:attentionMechanism}), and the attention coefficient in the masked attention mechanism represents the importance of a neighboring node $j \in \mathcal{N}_{i}$ to the central node $i$, where $\mathcal{N}_{i}$ is the set of all first-order neighbors of node $i$ (including $i$ itself).

The application of the multi-head attention mechanism (\textit{multi-head attention}) in \ref{headAttention} to graphs is described as follows:

\begin{equation}
	\label{maskAttention}
	\centering
	{e_{ij}}={f_{\text{mask attention}}(\mathbf{W} \overrightarrow{e_i}, \mathbf{W} \overrightarrow{e_j})}
\end{equation}

where $e_{ij}$ denotes the multi-head attention coefficient of an edge $(e_i, e_j)$ with respect to the central entity $e_i$ in the knowledge graph $\mathcal{G}_{\text{know}}$. $\mathbf{W}$ is a weight matrix that parameterizes the linear transformation. $f_{\text{mask attention}}$ is the function applying the attention mechanism.

In the GAT model, each entity vector embedding $\overrightarrow{e_i}$ undergoes two transformation stages. The entire model consists of two transformation steps, each applying the multi-head attention mechanism as follows:

\begin{equation}
	\label{gatProcess}
	\overrightarrow{e_i} \xrightarrow{f_{\text{mask attention}}^{(1)}} \overrightarrow{e'_i} \xrightarrow{f_{\text{mask attention}}^{(2)}} \overrightarrow{e''_i}
\end{equation}

In the first multi-head attention step ($f_{\text{mask attention}}^{(1)}$), the model aggregates information from neighboring entities and stacks them to produce vector $\overrightarrow{e'_i}$, where $\overrightarrow{e'_i} \in \mathbb{R}^{1 \times D'}$. In the second step ($f_{\text{mask attention}}^{(2)}$), the multi-head attention layer is no longer sensitive to self-attention; therefore, the output is computed as an \textit{average} instead of concatenating attention heads. The vector $\overrightarrow{e'_i}$ is then treated as the input embedding to be transformed into the final output embedding vector $\overrightarrow{e''_i}$, with $\overrightarrow{e''_i} \in \mathbb{R}^{1 \times D''}$.

First, similar to the attention mechanism in \ref{attention}, each embedding vector is multiplied by a weight matrix $\mathbf{W}_1 \in \mathbb{R}^{D_k \times D_{\text{in}}}$ to parameterize the linear transformation from $D_{\text{in}}$ input dimensions to $D_k$ higher-level feature dimensions:

\begin{equation}
	\overrightarrow{h_i} = \mathbf{W}_{1} \overrightarrow{e_i}
\end{equation}

where $\overrightarrow{e_i} \in \mathbb{R}^{D_{\text{in}} \times 1}
\xrightarrow{} \overrightarrow{h_i} \in \mathbb{R}^{D_k \times 1}$

Next, we concatenate each pair of linearly transformed entity embedding vectors to compute the attention coefficients. The attention coefficient $e_{ij}$ reflects the importance of the edge feature $(e_i, e_j)$ with respect to the central entity $e_i$, or in other words, the importance of a neighboring entity $e_j$ that is connected to $e_i$. We apply the $\text{LeakyReLU}$ function to extract the absolute value of the attention coefficient. Each attention coefficient $e_{ij}$ is computed using the following equation:

\begin{equation}
	e_{ij} = \Big( \text{LeakyReLU} \Big( \overrightarrow{\mathbf{W}_{2}}^{T} [\overrightarrow{h_i} || \overrightarrow{h_j}]\Big) \Big)
\end{equation}

where ${.}^{T}$ denotes the transpose operation and $||$ represents concatenation. This is similar to \ref{attention}, however instead of using a dot product, we use a \textit{shared attentional mechanism} $\overrightarrow{\mathbf{W}_2}$: $\mathbb{R}^{D_k} \times \mathbb{R}^{D_k} \rightarrow \mathbb{R}$ to compute the attention scores. As mentioned in \ref{maskAttention}, we perform self-attention between all nodes using the masked attention mechanism to discard all irrelevant structural information.

To enable meaningful comparison between the attention coefficients of neighboring entities, a \textit{softmax} function is applied to normalize the coefficients over all neighbors $e_j$ that are connected to the central entity $e_i$: $\alpha_{ij} = \text{softmax}_j(e_{ij})$. Combining all of this, we obtain the final normalized attention coefficient of each neighbor with respect to the central entity as follows:

\begin{equation}
	\label{attentionCoeff}
	\alpha_{ij} = \frac{
		\text{exp} \Big( \text{LeakyReLU} \Big( \overrightarrow{\mathbf{W}_2}^{T} [ \overrightarrow{h_i} || \overrightarrow{h_j}]\Big) \Big))
	}
	{
		\sum_{k \in \mathcal{N}_i}
		\text{exp} \Big( \text{LeakyReLU} \Big( \overrightarrow{\mathbf{W}_2}^{T} [\overrightarrow{h_i} || \overrightarrow{h_k}]\Big) \Big))
	}
\end{equation}

At this stage, the GAT model operates similarly to GCN \cite{kipf2016semi}, where the embedding vectors from neighboring nodes are aggregated and scaled by their corresponding normalized attention coefficients:

\begin{equation}
	\label{scaleAttentionCoef}
	\centering
	{\overrightarrow{e'_i}}={\sigma\left(\sum_{j\in \mathcal{N}_i} {\alpha_{ij} \overrightarrow{h_j} }\right)}
\end{equation}

Similar to the multi-head attention layer, we concatenate $N_{\text{head}}$ attention heads to stabilize the training process in the first step ($f_{\text{mask attention}}^{(1)}$ \ref{gatProcess}) of the model:

\begin{equation}
	\label{multiHeadAttention}
	{\overrightarrow{e'_i}}={\bigparallel_{h=1}^{N_{\text{head}}}\sigma\left(\sum_{j\in \mathcal{N}_i}\alpha_{ij}^{h} \mathbf{W}^{h} \overrightarrow{h_{j}} \right)}
\end{equation}

where $\sigma$ is any non-linear activation function, and $\alpha_{ij}^h$ is the normalized attention coefficient for edge $(e_i, e_j)$ computed from the $h^{th}$ attention head. Similar to equation \ref{attention}, $\mathbf{W}^h$ is the weight matrix used for linear transformation of the input embedding vector, with each $\mathbf{W}^h$ corresponding to a different attention head. The resulting new embedding vector $\overrightarrow{e'_i} \in \mathbb{R}^{1 \times D'}$, where $D' = N_{\text{head}} D_{\text{k}}$, is then used as the input for the next attention layer. However, in the second step ($f_{\text{mask attention}}^{(2)}$ \ref{gatProcess}), the multi-head attention outputs are averaged instead of concatenated, as shown below:

\begin{equation}
	\label{multiHeadConcat}
	{\overrightarrow{e''_i}}={\sigma\left(\frac{1}{N_{\text{head}}} \sum_{h=1}^{N_{\text{head}}}\sum_{j\in \mathcal{N}_i}\alpha_{ij}^{h} \mathbf{W}^{h} \overrightarrow{e'_{j}} \right)}
\end{equation}

\textbf{Summary}: Up to this point, we have presented how the attention mechanism aggregates a knowledge graph entity embedding vector from its neighboring embeddings and concatenates them to produce the final embedding. In the next section, we will present our complete embedding model based on the KBGAT model proposed by Nathani, Deepak\cite{nathani2019learning}.

\subsection{KBGAT Model}

In a knowledge graph, a single entity cannot fully represent an edge, as the entity can play multiple roles depending on the type of relation. For example, in \autoref{fig:graphExample}, Donald Trump serves both as a president and as a husband. To address this issue, the knowledge graph attention-based embedding model — KBGAT (graph attention based embeddings \cite{nathani2019learning}) — improves upon the GAT model by incorporating additional information from \textit{relations and neighboring node features} into the attention mechanism. In this section, we will detail the KBGAT model. The structure of KBGAT follows an encoder-decoder framework, where the encoder is implemented using the Graph Attention Network (GAT), and the decoder uses the ConvKB model for prediction. The steps of the KBGAT model are illustrated in \ref{eq:KBGATProcess}.

\begin{equation*}
	\label{eq:KBGATProcess}
	\resizebox{\linewidth}{!}{$
		entities \xrightarrow{\text{Embedding}^{\ref{sec:initTransE}}} \overrightarrow{e_{\text{TransE}}} \xrightarrow{\text{Embedding}^{\ref{sec:encodeKBGAT}}} \overrightarrow{e_{\text{KBGAT}}} \xrightarrow{\text{ConvKB}^{\ref{sec:predictionConvKB}}} e_{\text{prob}}
		$}
\end{equation*}

First, the embedding vectors of each entity are initialized using the TransE model to capture the spatial characteristics among nodes and obtain the initial embeddings. These embeddings are then further trained using an encoder model to capture neighborhood features, resulting in updated embeddings. Finally, these embeddings are passed through a prediction layer using the ConvKB model. All equations presented here are based on those in the work of Nathani, Deepak\cite{nathani2019learning}.

\subsubsection{Embedding Initialization}
\label{sec:initTransE}

\begin{figure}[htp]
	\centering
	\resizebox{\linewidth}{!}{
	
	\begin{tikzpicture}[>=Stealth,
	arrow/.style={->,thick},
	vector/.style={arrow, ultra thick},
	dashLine/.style={->, thin, dash pattern=on 1mm off 0.5mm},
	formal/.style={font=\sffamily}]
	\coordinate (root) at (0,0);
	\coordinate [formal, label=above left:$\overrightarrow{\text{relation}}$] (c1) at (-1.5, 3);
	\coordinate [formal, label=above:$\overrightarrow{\text{tail}_{\text{invalid}}}$] (c2) at (3.5, 5);
	\coordinate [formal, label=right:$\overrightarrow{\text{tail}_{\text{valid}}}$] (c3) at (3.5, 4);
	\coordinate [formal, label=above right :$\overrightarrow{\text{head}_{\text{invalid}}}$] (c4) at (5, 2);
	\coordinate [formal, label=right:$\overrightarrow{\text{head}_{\text{valid}}}$] (c5) at (5, 1);
	
	\draw [arrow] ([yshift=-2em] root) -> (0, 5) node (yaxis)[above] {$y$};
	\draw [arrow] ([xshift=-5em] root) -> (5, 0) node (xaxis)[above] {$x$};
	
	\foreach \x in {3, 5}
	{\draw [vector, color=azure] (root) -> (c\x);}
	
	\foreach \x in {2, 4}
	{\draw [vector, color=awesome] (root) -> (c\x);}
	
	\draw [vector, color=amber] (root) -> (c1);
	
	\foreach \x in {1, 4}
	{\draw [dashLine, color=darkpastelgreen] (c\x) -> (c2);}
	
	\foreach \x in {1, 5}
	{\draw [dashLine, color=darkpastelgreen] (c\x) -> (c3);}
\end{tikzpicture}
	}
	\caption{Illustration of embedding vectors in the TransE model}
	\label{fig:TransEAnimation}
\end{figure}
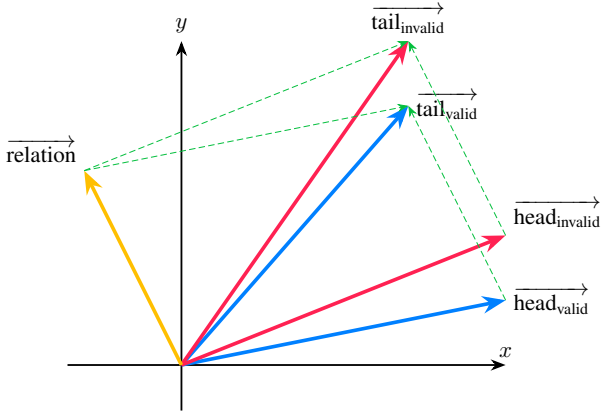

Similar to the Word2Vec method \cite{mikolov2013efficient}, the model \textit{Translating Embeddings for Modeling Multi-relational Data} (TransE \cite{bordes2013translating}) belongs to the group of geometric embedding methods that transform entities and relations in a knowledge graph into output embedding vectors such that:
\begin{equation}
	\label{eq:conditionTransE}
	\overrightarrow{\text{entity}_{\text{head}}} + \overrightarrow{relation} \approx \overrightarrow{\text{entity}_{\text{tail}}}
\end{equation}

Initially, the entity and relation embedding vectors are randomly initialized using a normal distribution with dimensionality $D_{\text{in}}$, and then normalized according to the size of the entity and relation embedding sets.

\begin{figure*}[htbp]
	\centering
	\resizebox{0.9\linewidth}{!}{%
	\begin{subfigure}[b]{0.45\textwidth}
		\centering
		\begin{tikzpicture}[>=Stealth, scale=0.7]
	\draw[step=1cm, thin, gray!60] (-5,-1) grid (5,9);
	\draw[very thick] (-5,0) -- (5,0);    
	\draw[very thick] (0,-1) -- (0,9);    
	
	\tikzset{
		point/.style = {circle, fill=azure, inner sep=2pt},
		vblue/.style  = {thick, amber},
		vblack/.style = {thick, black},
		vred/.style   = {thick, awesome},
		vgreen/.style = {thick, darkpastelgreen},
	}
	
	\node[point,label=below left:A] (A) at (1,1) {};
	\node[point,label=above right:B] (B) at (3,4) {};
	\draw[vblue,->] (A) -- (B) node[midway, below right=2pt] {\small$h=(2,3)$};
	
	\node[point,label=left:C]  (C) at (1,3) {};
	\node[point,label=left:D]  (D) at (1,5) {};
	\draw[vblack,->] (C) -- (D) node[midway, left=4pt] {\small$r=(0,2)$};
	
	\node[point,label=above left:E] (E) at (2,6) {};
	\node[point,label=above right:F] (F) at (4,7) {};
	\draw[vred,->] (E) -- (F) node[midway, above=2pt, sloped] {\small$t'=(2,1)$};
	
	\node[point,label=right:G] (G) at (-2,5) {};
	\node[point,label=above left:H] (H) at (-3,7) {};
	\draw[vgreen,->] (G) -- (H) node[midway, right=2pt] {\small$t=(-1,2)$};
	
\end{tikzpicture}
		\caption{Initial TransE embedding vector}
		\label{fig:TransEEmbeddingBefore}
	\end{subfigure}
	\hfill
	\begin{subfigure}[b]{0.45\textwidth}
		\centering
		\begin{tikzpicture}[>=Stealth, scale=0.7]
	\draw[step=1cm, thin, gray!60] (-5,-1) grid (5,9);
	\draw[very thick] (-5,0) -- (5,0);    
	\draw[very thick] (0,-1) -- (0,9);    
	
	\tikzset{
		point/.style = {circle, fill=azure, inner sep=2pt},
		vblue/.style  = {thick, amber},
		vblack/.style = {thick, black},
		vred/.style   = {thick, awesome},
		vgreen/.style = {thick, darkpastelgreen},
		vdash/.style  = {thick, dashed, deepmagenta},
		vline/.style  = {thick, dashed, gray!80}
	}
	
	\node[point,label=below right:A] (A) at (1,3) {};
	\node[point,label=right:B] (B) at (3,6	) {};
	\draw[vblue,->] (A) -- (B) node[midway, below right=2pt] {\small$h=(2,3)$};
	
	\node[point,label=below left:C] (C) at (1,3) {};
	\node[point,label=left:D] (D) at (1,5) {};
	\draw[vblack,->] (C) -- (D) node[midway, left=4pt] {\small$r=(0,2)$};
	
	\node[point,label=above left:E] (E) at (2,6) {};
	\node[point,label=above right:F] (F) at (4,7) {};
	\draw[vred,->] (E) -- (F) node[midway, above=10pt, sloped] {\small$t'=(2,1)$};
	
	\node[point,label=right:G] (G) at (-2,5) {};
	\node[point,label=above:H] (H) at (-3,7) {};
	\draw[vgreen,->] (G) -- (H) node[midway, right=2pt] {\small$t=(-1,2)$};
	
	\node[point,label=left:K] (K) at (-2.5,6) {};
	\node[point,label=above:I] (I) at (2.45,6.2) {};
	\node[point,label=below:J] (J) at (2.6,5.4) {};
	
	\draw[vline] (K) -- (J) node[midway, above=0pt, xshift=3mm] {\small$\color{deepmagenta}{d}$};
	\draw[vline] (J) -- (I) node[midway, xshift=-2mm] {\small$\color{deepmagenta}{d'}$};
	
\end{tikzpicture}
		
		\caption{Embedding after transformation}
		\label{fig:TransEEmbeddingAfter}
	\end{subfigure}
	}
	\caption{The TransE graph embedding vector}
	\label{fig:TransEExplain}
\end{figure*}
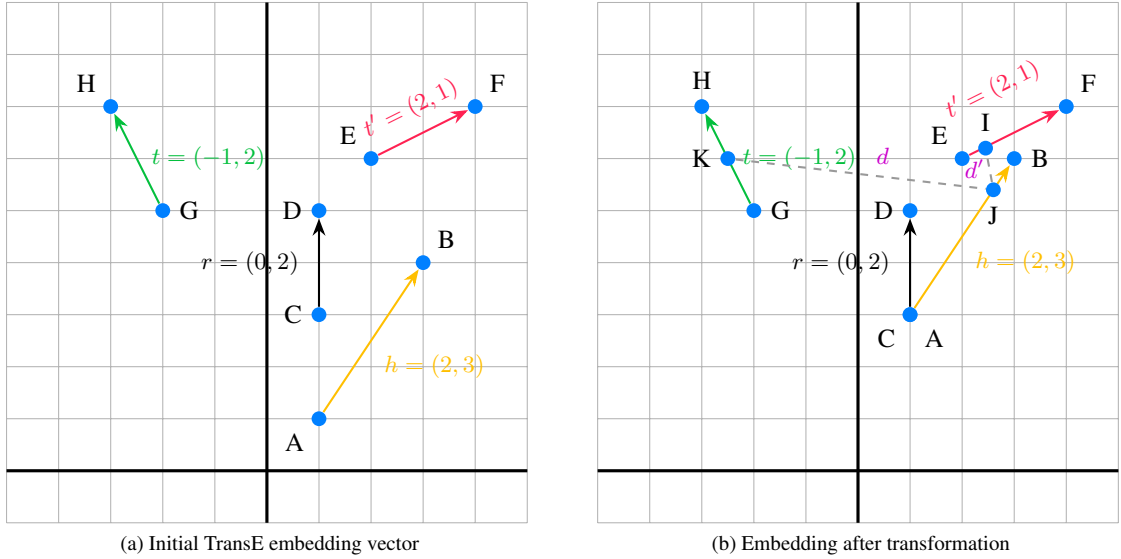

Next, we perform sampling from the training dataset to obtain a batch of valid triples ($S_{\text{batch}}$). For each such triple, we sample an invalid triple by replacing either the head or the tail entity with a random entity from the entity set, yielding a batch of invalid triples ($S'_{\text{batch}}$). We then pair each valid triple with an invalid one to form the training batch ($T_{\text{batch}}$). Finally, we update the embedding vectors to satisfy the condition in \ref{eq:conditionTransE}.

\begin{algorithm}
	\caption{TransE Embedding Learning Algorithm \protect\cite{bordes2013translating}}\label{alg:TransE}
	\begin{algorithmic}[1]
		\Statex \textbf{Input} :
		Training set $S = {(h, r, t)}$, entity set $E$, relation set $R$, margin $\gamma$, embedding dimension $D_{\text{in}}$	
		\Statex \textbf{Initialize}
		\State $\overrightarrow{r} \leftarrow \text{uniform}(-\frac{6}{\sqrt{D_{\text{in}}}}, \frac{6}{\sqrt{D_{\text{in}}}})$ for each relation $r \in R$
		\State $\overrightarrow{r} \leftarrow \frac{\overrightarrow{r}}{\|\overrightarrow{r}\|}$ for each $r \in R$
		\State $\overrightarrow{e} \leftarrow \text{uniform}(-\frac{6}{\sqrt{D_{\text{in}}}}, \frac{6}{\sqrt{D_{\text{in}}}})$ for each entity $e \in E$
		\Loop
		\State $\overrightarrow{e} \leftarrow \frac{\overrightarrow{e}}{\|\overrightarrow{e}\|}$ for each $e \in E$
		\State $S_{\text{batch}} \leftarrow \text{sample}(S, b)$  // sample minibatch of size $b$
		\State $T_{\text{batch}} \leftarrow \varnothing $
		\For {$(h, r, t) \in S_{\text{batch}}$}
		\State $(h', r, t') \leftarrow \text{sample}(S'_{(h, r, t)})$ // sample from invalid triple set
		\State $T_{\text{batch}} \leftarrow T_{\text{batch}} \cup \Big\{ \Big( (h, r, t), (h', r, t') \Big) \Big\}$
		\EndFor
		\Statex Update embeddings
		\State $\sum_{\Big( (h, r, t), (h', r, t')\Big) \in T_{\text{batch}}} \nabla [\gamma + d(\overrightarrow{h} + \overrightarrow{r}, \overrightarrow{t}) - d(\overrightarrow{h'} + \overrightarrow{r}, \overrightarrow{t'})]_{+}$
		\EndLoop
		\Statex \textbf{Output} :
		A set of embedding vectors with dimension $D_{\text{in}}$ representing entities and relations
	\end{algorithmic}
\end{algorithm}

The TransE model proposed by Bordes, Antoine \cite{bordes2013translating} is presented in Algorithm \ref{alg:TransE}.

The input of the TransE model is a training dataset where each element is a triple $(h, r, t)$. Here, $h, t \in E$ are the head and tail entities, and $r \in R$ is the relation. $\overrightarrow{e}$ and $\overrightarrow{r}$ are the embedding vectors of entities and relations respectively, and $\|\overrightarrow{e}\|$ and $\|\overrightarrow{r}\|$ denote the cardinalities of the entity and relation sets. $S$ and $S_{\text{batch}}$ represent the full training dataset and a sampled batch from it, respectively. $T_{\text{batch}}$ is a batch containing both valid and invalid triples used to compute the loss function \ref{eq:sampleTransE}.

A \textit{valid triple} is one directly sampled from the training set ($S_{\text{batch}}$), while an \textit{invalid triple} is constructed by corrupting a valid triple ($S'_{\text{batch}}$) by replacing either the head or the tail entity with a randomly selected entity from the entity set:

\begin{equation}
	\label{eq:sampleTransE}
	\centering
	S'_{(h, r, t)} = \big\{ (h', r, t) | h' \in E \big\} \cup \big\{ (h, r, t') | t' \in E \big\}
\end{equation}

To achieve the goal of learning embedding vectors such that $\overrightarrow{h} + \overrightarrow{r} \approx \overrightarrow{t}$, the model aims for the tail embedding $\overrightarrow{t}$ of valid triples to lie close to $\overrightarrow{h} + \overrightarrow{r}$, while in invalid triples, the corrupted embedding $\overrightarrow{h'} + \overrightarrow{r}$ (or $\overrightarrow{t'}$) should lie far from $\overrightarrow{t}$ (or $\overrightarrow{h} + \overrightarrow{r}$), according to the following margin-based ranking loss function:

\begin{equation}
	\label{eq:KBGATLoss}
	\centering
	\mathcal{L} = \sum_{(h, r, t) \in S} \sum_{(h', r, t') \in S'_{(h, r, t)}} [d - d' + \gamma]_{+}
\end{equation}

Here, $\gamma > 0$ is the margin, and $h'$ and $t'$ are entities sampled as defined in Equation \ref{eq:sampleTransE}. $\Delta$ and $\Delta'$ represent the difference vectors for the embeddings in valid and invalid triples, respectively, with $d = \big\|\Delta \big\|_{1} = \big\| \overrightarrow{h} + \overrightarrow{r} - \overrightarrow{t} \big\|_{1}$ and $d' = \big\|\Delta' \big\|_{1} = \big\| \overrightarrow{h'} + \overrightarrow{r} - \overrightarrow{t'} \big\|_{1}$, where $\|\cdot\|_{1}$ denotes the L1 norm.

As illustrated in \autoref{fig:TransEEmbeddingBefore}, \autoref{fig:TransEEmbeddingAfter}, if $d > d'$ or $d - d' > 0$, then $\overrightarrow{h} + \overrightarrow{r}$ is closer to $\overrightarrow{t}$ than to $\overrightarrow{t'}$. Since we want the embedding vectors to satisfy the condition in Equation \ref{eq:conditionTransE}, $\overrightarrow{h} + \overrightarrow{r}$ should be as close as possible to $\overrightarrow{t}$. That means, the closer $\overrightarrow{h} + \overrightarrow{r}$ is to $\overrightarrow{t'}$, the more incorrect it becomes.

Therefore, during training, we aim for $\Delta'$ to be as large as possible relative to $\Delta$. If $\Delta' > \Delta$ or $d' - d > 0$, there is no need to update the embedding weights further. Hence, in the loss function \ref{eq:KBGATLoss}, the term $[d - d' + \gamma]_{+}$ captures only the positive part because the negative part already satisfies the correctness of the condition in Equation \ref{eq:conditionTransE} during training.

\subsubsection{Encoder Model}
\label{sec:encodeKBGAT}

After obtaining the embedding vectors that capture spatial features of the knowledge graph, these embeddings are passed through the next embedding layer to further aggregate the neighborhood information of each entity.

The model transforms the entity embedding matrix

\[
\mathbf{E} = \Big\{\overrightarrow{e_1}, \overrightarrow{e_2}, ...,  \overrightarrow{e_{N_e}}\Big\} \xrightarrow{} \mathbf{E''} = \Big\{\overrightarrow{e''_1}, \overrightarrow{e''_2}, ...,  \overrightarrow{e''_{N_e}}\Big\},
\]
with $\mathbf{E} \in \mathbb{R}^{N_e \times D_{\text{in}}}$ and $\mathbf{E''} \in \mathbb{R}^{N_e \times D''}$.

Simultaneously, it transforms the relation embedding matrix

\[
\mathbf{R} = \Big\{\overrightarrow{r_1}, \overrightarrow{r_2}, ...,  \overrightarrow{r_{N_r}}\Big\} \xrightarrow{} \mathbf{R''} = \Big\{\overrightarrow{r''_1}, \overrightarrow{r''_2}, ...,  \overrightarrow{r''_{N_r}}\Big\},
\]
with $\mathbf{R} \in \mathbb{R}^{N_r \times P_{\text{in}}}$ and $\mathbf{R''} \in \mathbb{R}^{N_r \times P''}$.

Similar to the GAT model described in Section~\ref{sec:GAT}, the model transforms the entity embedding vectors from $D_{\text{in}}$ dimensions to $D''$ dimensions by aggregating neighborhood information through attention coefficients. $P_{\text{in}}$ and $P''$ are the input and output dimensions of the relation embedding vectors, respectively. $N_e$ and $N_r$ are the sizes of the entity and relation sets in $\mathcal{G}_{know}$, respectively.

\begin{figure*}[htp]
	\centering
	\resizebox{\textwidth}{!}{%
		\begin{tikzpicture}[
	emb/.style={draw, circle,minimum width=.1em, thin, anchor=center},
	entityInit/.style={emb, fill=amber},
	relationInit/.style={emb, fill=azure},
	entityEmb/.style={emb, fill=gray},
	relationEmb/.style={emb, fill=deepcarminepink},
	entityPretrain/.style={emb, fill=deepmagenta},
	entityOut/.style={emb, fill=deepskyblue},
	box/.style={draw,rectangle, fill=none},
	embBox/.style={box ,rounded corners=0.2em, minimum width=1.3em},
	layerBox/.style={color=azure!80, box ,very thick,rounded corners=0.2em, minimum width=1.3em},
	textbox/.style={box,rounded corners=0.3em, fill=none, align=center, execute at begin node=\setlength{\baselineskip}{1em}},
	arrowStrong/.style={thick,->,>=stealth},
	arrowDash/.style={thick, ->,>=stealth, dash pattern=on 1.5mm off 0.7mm, postaction={decorate}},
	row/.style={draw, rectangle, font=\fontsize{0.6em}{1}\sffamily, align=center, text width=3.9em},
	arrowStyle/.style={-latex', font=\sffamily},
	mydot/.style={draw, circle, minimum size=0.1em, scale=0.2},
	]
	\draw node[textbox][text width=5.7em] (graphAttention2) {\small Graph Attention Layer $2$};
	
	\draw node[textbox][minimum width=9.5em, minimum height=21em, left=1.5em of graphAttention2] (layer3) {};
	
	\draw node[embBox][minimum height=20.3em, left=2em of graphAttention2] (box10) {};
	\begin{scope}[every node/.style={below=0mm of box10.center}]
		\foreach \x in {1,...,12}
		{\draw node[entityEmb][yshift=(\x-4)*1.1em+0.55em] (ex10\x) {};}
		
		\foreach \x in {1,...,6}
		{\draw node[relationEmb](ey10\x)[yshift=-(\x+3)*1.1em+0.55em]{};}
	\end{scope}
	
	\draw node[embBox][minimum height=7em, left=2em of box10, yshift=5em] (box7) {};
	\draw node[embBox][minimum height=7em, left=2em of box10, yshift=-5em] (box9) {};
	\begin{scope}
		\foreach \x in {1,...,6}
		{\draw node[entityEmb](e7\x)[yshift=(\x-3)*1.1em, below=0mm of box7.center]{};}
		
		\foreach \x in {1,...,6}
		{\draw node[relationEmb](e9\x)[yshift=(\x-3)*1.1em, below=0mm of box9.center]{};}
	\end{scope}
	
	\draw node[embBox][minimum height=4em, left=2em of box7, yshift=2.5em] (box5) {};
	\draw node[embBox][minimum height=4em, left=2em of box7, yshift=-2.5em] (box6) {};
	\draw node[embBox][minimum height=5em, left=2em of box9] (box8) {};
	\begin{scope}
		\foreach \x in {1,...,3}
		{\draw node[entityEmb](e5\x)[yshift=(\x-2)*1.1em+0.55em, below=0mm of box5.center]{};}
		
		\foreach \x in {1,...,3}
		{\draw node[entityEmb](e6\x)[yshift=-(\x-2)*1.1em+0.55em, below=0mm of box6.center]{};}
		
		\foreach \x in {1,...,4}
		{\draw node[relationInit](e8\x)[yshift=(\x-2)*1.1em, below=0mm of box8.center]{};}
	\end{scope}
	\draw node[layerBox][minimum height=8em,minimum width=5.5em, left=1.5em of box10, yshift=-5em] (fc1) {};
	
	\begin{scope}[every node/.style={left=2em of layer3}]
		\draw node[textbox][yshift=5.5em, text width=4em] (attentionHead1) {\small Attention Head 1};
		\draw node[textbox][yshift=-5.5em, text width=4em] (attentionHead2) {\small Attention Head 2};
	\end{scope}
	\draw node[textbox][left=1.5em of layer3, text width=5.5em, minimum width=5.8em, minimum height=16em] (graphAttention1) {\small Graph Attention Layer $1$ };
	
	\draw node[textbox][minimum width=13em, minimum height=21em, left=1.5em of graphAttention1] (layer1) {};
	\draw node[embBox][minimum height=10.5em, yshift=1em,right=3em of layer1.center, left=2em of graphAttention1] (box4) {};
	\begin{scope}
		\foreach \x in {1,...,6}
		{\draw node[entityInit](ex4\x)[yshift=(\x-2)*1.1em-0.4em, above=0em of box4.center]{};}
		
		\foreach \x in {1,...,3}
		{\draw node[relationInit](e4y\x)[yshift=-(\x)*1.1em-0.6em, below=0mm of box4.center]{};}
	\end{scope}
	
	\draw node[embBox][minimum height=4em, yshift=4.5em, left=2em of box4] (box1) {};
	\foreach \x in {1,...,3}
	{\draw node[entityInit](e1\x)[yshift=(\x-2)*1.1em+0.5em, below=0mm of box1.center]{};}
	
	\draw node[embBox][minimum height=4em, left=2em of box4] (box2) {};
	\foreach \x in {1,...,3}
	{\draw node[relationInit](e2\x)[yshift=(\x-2)*1.1em+0.5em, below=0mm of box2.center]{};}
	
	\draw node[embBox][minimum height=4em, yshift=-4.5em, left=2em of box4] (box3) {};
	\foreach \x in {1,...,3}
	{\draw node[entityInit](e3\x)[yshift=(\x-2)*1.1em+0.5em, below=0mm of box3.center]{};}
	
	\begin{scope}[every node/.style={left=of box2, row}]
		\draw node (relation1) {richest\_of};
		\draw node[above=0mm of relation1] (entity1) {Elon Musk};
		\draw node[above=0mm of entity1] (triple1) {Triple 1};
		\draw node[below=0mm of relation1] (entity2) {United States};	
	\end{scope}
	
	\begin{scope}[every node/.style={row}]
		\draw node[below=3em of entity2] (triple2) {Triple N};
		\draw node[below=0mm of triple2] (entity3) {Tom Cruise};
		\draw node[below=0mm of entity3] (relation2) {born\_in};
		\draw node[below=0mm of relation2] (entity4) {New York};	
	\end{scope}
	
	\draw node[textbox][minimum height=16em, minimum width=2.3em, yshift=-4em, right=1.5em of graphAttention2] (layer4){};
	\draw node[embBox][minimum height=7em, right=2em of graphAttention2] (box11) {};
	\begin{scope}
		\foreach \x in {1,...,6}
		{\draw node[entityEmb](e11\x)[yshift=(\x-3)*1.1em, below=0mm of box11.center]{};}
	\end{scope}
	
	\draw node[embBox][minimum height=7em, below=0.7em of box11] (box12) {};
	\begin{scope}
		\foreach \x in {1,...,6}
		{\draw node[relationEmb](e12\x)[yshift=(\x-3)*1.1em, below=0mm of box12.center]{};}
	\end{scope}
	
	\draw node[embBox][minimum height=7em, above=1.5em of box11] (box13) {};
	\begin{scope}
		\foreach \x in {1,...,6}
		{\draw node[entityPretrain](e13\x)[yshift=(\x-3)*1.1em, below=0mm of box13.center]{};}
	\end{scope}
	
	\draw node[embBox][minimum height=5em, left=1.7em of box13] (box14) {};
	\begin{scope}
		\foreach \x in {1,...,4}
		{\draw node[entityInit](e14\x)[yshift=(\x-2)*1.1em, below=0mm of box14.center]{};}
	\end{scope}
	\draw node[layerBox][minimum height=8em, minimum width=5.5em, above=1.5em of box11, yshift=-0.5em, xshift=-1.5em] (fc2) {};
	
	\draw node[right=2em of box11, yshift=3.7em] (bigOTimes) {$\bigotimes$};
	
	\draw node[textbox][minimum height=16em, minimum width=2.3em, yshift=-3.95em, right=1.5em of bigOTimes] (layer5) {};
	\draw node[embBox][minimum height=7em, above=1em of layer5.center, yshift=-0.5em] (box15) {};
	\begin{scope}
		\foreach \x in {1,...,6}
		{\draw node[entityOut](e15\x)[yshift=(\x-3)*1.1em, below=0mm of box15.center]{};}
	\end{scope}
	
	\draw node[embBox][minimum height=7em, below=1em of layer5.center, yshift=0.5em] (box16) {};
	\begin{scope}
		\foreach \x in {1,...,6}
		{\draw node[relationEmb](e16\x)[yshift=(\x-3)*1.1em, below=0mm of box16.center]{};}
	\end{scope}
	
	\draw node[textbox][right=1.5em of layer5] (final) {$\mathcal{L}$};
	\draw [arrowStrong] (box10) -- (graphAttention2.west);
	
	\draw [arrowStrong] (entity1.east) -- (box1.west);
	\draw [arrowStrong] (relation1) -- (box2.west);
	\draw [arrowStrong] (entity2.east) -- (box3.west);
	
	\draw [arrowStrong] (graphAttention2.east) -- (box11.west);
	\draw [arrowStrong] (graphAttention2.east) -- (box12.west);
	
	\draw [arrowStrong] (box11.east) to ([xshift=0.35em] bigOTimes.west);
	\draw [arrowStrong] (box13.east) to ([xshift=0.35em] bigOTimes.west);
	
	\draw [arrowStrong] ([xshift=-1.5mm] bigOTimes.east) -- (box15.west);
	
	\draw [arrowStrong] (box16.east) -- (final.west);
	\draw [arrowStrong] (box15.east) -- (final.west);
	
	\foreach \x in {1,2}
	\draw [arrowStrong] (layer1.east) -- (attentionHead\x.west);
	
	\foreach \x in {1,2}
	\draw [arrowStrong] (attentionHead\x.east) -- (layer3.west);
	
	\foreach \x in {1,2,3}
	\draw [arrowDash] (box\x.east) -- (box4.west);
	
	\foreach \x in {7,9}
	{\draw [arrowDash] (box\x) -- (box10.west);}
	
	\foreach \x in {5,6}
	{\draw [arrowDash] (box\x) -- (box7.west);}
	
	\foreach \x in {1, ..., 6}
	{\draw [thick] (e81) -- (e9\x);}
	\foreach \x in {1, ..., 6}
	{\draw [thick] (e82) -- (e9\x);}
	\foreach \x in {1, ..., 6}
	{\draw [thick] (e83) -- (e9\x);}
	\foreach \x in {1, ..., 6}
	{\draw [thick] (e84) -- (e9\x);}
	
	\foreach \x in {1, ..., 6}
	{\draw [thick] (e141) -- (e13\x);}
	\foreach \x in {1, ..., 6}
	{\draw [thick] (e142) -- (e13\x);}
	\foreach \x in {1, ..., 6}
	{\draw [thick] (e143) -- (e13\x);}
	\foreach \x in {1, ..., 6}
	{\draw [thick] (e144) -- (e13\x);}
	
	\draw node[mydot][below=1.1em of entity2] (dot1) {};
	\draw node[mydot][below=1.4em of entity2] (dot2) {};
	\draw node[mydot][below=1.7em of entity2] (dot3) {};
\end{tikzpicture}
	}
	\caption{Illustration of the encoder layers in the GCAT model}
	\label{fig:encoderKBGAT}
\end{figure*}
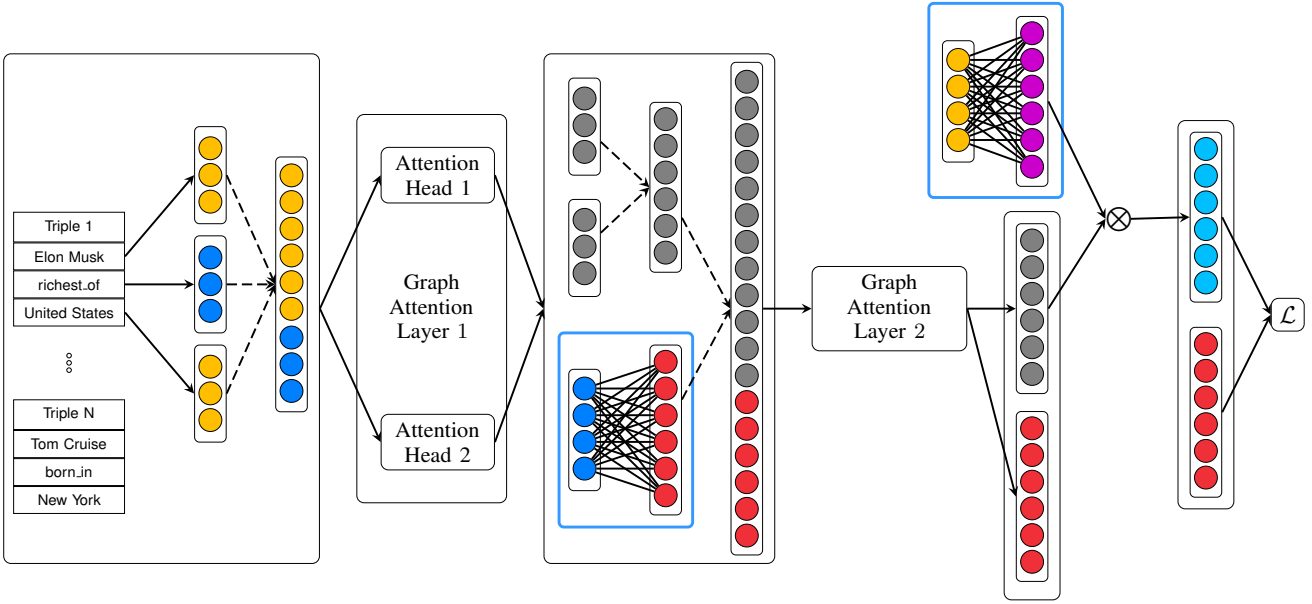

The KBGAT model concatenates the entity and relation embedding vectors according to the following structure:

\begin{equation}
	\label{attentionWithRelation}
	\overrightarrow{t_{ijk}} = \mathbf{W_1} [\overrightarrow{e_i} || \overrightarrow{e_j} || \overrightarrow{r_k}]
\end{equation}

Here, $\overrightarrow{t_{ijk}}$ is the embedding vector representing the triple $t_{ij}^k = (e_i, r_k, e_j)$, where $e_j$ and $r_k$ are the neighboring entity and the relation connecting the source node $e_i$ to the node $e_j$. $\mathbf{W_1} \in \mathbb{R}^{D_k \times (2 D_{\text{in}} + P_{\text{in}})}$ is a weight matrix that performs a linear transformation of the concatenated input vectors into a new vector with dimensionality $D_k$. These weight matrices are either randomly initialized using a normal distribution or pre-trained using the TransE model~\cite{bordes2013translating}.

Similar to Equation~\ref{attentionCoeff} in the GAT model, we need to compute the attention coefficient for each edge with respect to a given node. Then, the \textit{softmax} function is applied to normalize these coefficients as follows:

\begin{equation}
	\label{attentionRelationCoeff}
	\begin{split}
		\alpha_{ijk}& = \text{softmax}_{jk}(\text{LeakyReLU}(\mathbf{W_2} \overrightarrow{t_{ijk}}))\\
		&= \frac{
			\text{exp} \Big( \text{LeakyReLU} \Big( \mathbf{W_2} \overrightarrow{t_{ijk}}\Big) \Big)
		}
		{
			\sum_{n\in \mathcal{N}_i} \sum_{r\in \mathcal{R}_{in}}
			\text{exp} \Big( \text{LeakyReLU} \Big( \mathbf{W_2} \overrightarrow{t_{inr}} \Big) \Big)
		}
	\end{split}
\end{equation}

where $\mathcal{N}_i$ denotes the set of neighbors of the central node $e_i$ within $n_{\text{hop}}$ hops; $\mathcal{R}_{in}$ represents the set of all relations that exist along the paths connecting the source entity $e_i$ to a neighboring entity $e_n \in \mathcal{N}_i$. Similar to Equation~\ref{scaleAttentionCoef}, the embedding vectors $\overrightarrow{t^k_{ij}}$ are scaled by their corresponding normalized attention coefficients:

\begin{align}
	{\overrightarrow{e'_{i}}}&={\sigma\left(\sum_{j \in \mathcal{N}_i} \sum_{k \in \mathcal{R}_{ij}} \alpha_{ijk} \overrightarrow{t_{ijk}}\right)}
\end{align}

Similar to the multi-head attention mechanism in Equation~\ref{multiHeadAttention}, we concatenate $N_{\text{head}}$ attention heads to stabilize the learning process:

\begin{equation}
	\label{eq:graphAttention1}
	\overrightarrow{e'}_i=\bigparallel_{h=1}^{N_{\text{head}}} \sigma\left(\sum_{j\in \mathcal{N}_i}\alpha_{ijk}^{(h)} \overrightarrow{t^{(h)}_{ijk}}\right)
\end{equation}

Likewise, for relation embeddings, a weight matrix $\mathbf{W}_R$ is used to perform a linear transformation from the original relation embedding of dimension $P$ to a new dimension $P'$:

\begin{align}
	\mathbf{R'} = \mathbf{R} \mathbf{W}^R; \hspace{2cm} \text{where: } \mathbf{W}^R \in \mathbb{R}^{P \times P'}
\end{align}

At this stage, we have obtained two matrices: $\mathbf{H}' \in \mathbb{R}^{N_e \times D'}$ and $\mathbf{R}' \in \mathbb{R}^{N_r \times P'}$, which are the updated entity and relation embedding matrices, respectively, with new dimensions. The model proceeds through the final attention layer, taking as input the newly updated entity and relation embeddings as shown in Equation~\ref{multiHeadConcat}. However, if we apply multi-head attention at this final layer for prediction, the concatenation operation will no longer be \textit{sensitive} to the self-attention mechanism. Therefore, instead of concatenating, the model averages the outputs and then applies a final non-linear activation function:

\begin{equation}
	\label{eq:graphAttention2}
	\overrightarrow{e''_{i}}=\sigma\left(\frac{1}{N_{\text{head}}} \sum_{h=1}^{N_{\text{head}}} \sum_{j \in \mathcal{N}_i} \sum_{k \in \mathcal{R}_{ij}} \alpha'^{(h)}_{ijk} \overrightarrow{t'^{(h)}_{ijk}} \right)
\end{equation}

where $\alpha'^{(h)}_{ijk}$ and $t'^{(h)}_{ijk}$ denote the normalized attention coefficients and the triple embedding vectors for $(e_i, r_k, e_j)$ in attention head $(h)$, respectively.

Up to this point, the KBGAT model functions similarly to the GAT model in Section~\ref{sec:GAT}, but it additionally incorporates both entity embedding information and neighbor nodes up to $n_{\text{hop}}$ hops. This results in the final entity embedding matrix $\mathbf{E}'' \in \mathbb{R}^{N_e \times D''}$ and the final relation embedding matrix $\mathbf{R}'' \in \mathbb{R}^{N_r \times P''}$. 

However, after the embedding learning process, the final entity embedding matrix $\mathbf{E}''$ may lose the initial embedding information due to the \textbf{vanishing gradient} problem. To address this, the model employs residual learning, by projecting the initial embedding matrix $\mathbf{E}$ through a weight matrix $\mathbf{W}^E \in \mathbb{R}^{D_{\text{in}} \times D''}$, and then directly adding it to the final embedding, thus preserving the initial embedding information during training:

\begin{equation}
	\label{eq:reInitEmbedding}
	\mathbf{H} = \mathbf{W}^E \mathbf{E} + \mathbf{E''}
\end{equation}

Finally, the training datasets are sampled to generate valid triples and invalid triples, similar to the TransE model described above, in order to learn the embedding vectors. However, the distance between embedding vectors is computed using the L1 norm as follows:
\[
d_{t_{ij}} = \big|\big|\vec{h_i}+ \vec{g_k}-\vec{h_j}\big|\big|_1
\]

Similarly, we train the model using a margin-based loss function:

\begin{equation}
	L(\Omega)=\sum_{t_{ij} \in S} \sum_{t'_{ij} \in S'} \text{max}\{d_{t'_{ij}} - d_{t_{ij}} + \gamma , 0 \}
\end{equation}

where $\gamma > 0$ is the margin parameter, $S$ is the set of valid triples, and $S'$ is the set of invalid triples, defined as:

\begin{equation}
	{S'} ={\underbrace{\{ t^k_{i'j} | e'_i \in \mathcal{E}\setminus e_i\}}_{\text{head entity replacement}} \cup \underbrace{\{ t^k_{ij'} | e'_j \in \mathcal{E}\setminus e_j\}}_{\text{tail entity replacement}}}
\end{equation}

The output of the GCAT model is the set of entity and relation embedding vectors, which are subsequently fed into the ConvKB model for final prediction.

\subsubsection{ConvKB Prediction Model}
\label{sec:predictionConvKB}

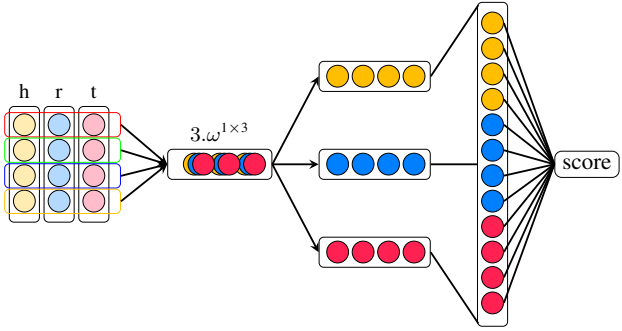
\begin{figure}[htp]
	\centering
	\resizebox{\linewidth}{!}{%
		\begin{tikzpicture}[
	emb/.style={draw, circle,minimum width=.1em, thin, anchor=center},
	head/.style={emb, fill=amber},
	relation/.style={emb, fill=azure},
	tail/.style={emb, fill=awesome},
	conv/.style={emb, fill=gray},
	box/.style={draw,rectangle, fill=none},
	embBox/.style={box ,rounded corners=0.2em},
	convBox/.style={box, minimum height=1.3em, minimum width=4.8em,rounded corners=0.2em},
	convolution/.style={box, minimum height=1.3em, minimum width=4.5em,rounded corners=0.2em},
	convLayer/.style={box, minimum height=1.05em, minimum width=5em, rounded corners=0.2em},
	textbox/.style={box,rounded corners=0.3em, fill=none, align=center},
	title/.style={fill=none, align=center},
	arrow/.style={thick,>=stealth},
	arrowStrong/.style={thick,->,>=stealth},
	arrowStyle/.style={-latex'},
	mydot/.style={draw, circle, minimum size=0.1em, scale=0.2},
	]
	\draw node[textbox] (lossConvKB) {score};
	
	\draw node[embBox][minimum width=1.3em, minimum height=14em, left=2em of lossConvKB] (box9) {};
	\foreach \x in {1,...,4}
	{\draw node[head](e9\x)[yshift=-(\x-7)*1.1em, below=0mm of box9.center]{};}
	
	\foreach \x in {1,...,4}
	{\draw node[relation](e8\x)[yshift=(\x-2)*1.1em, below=0mm of box9.center]{};}
	
	\foreach \x in {1,...,4}
	{\draw node[tail](e7\x)[yshift=(\x-6)*1.1em, below=0mm of box9.center]{};}

	\draw node[convBox][yshift=3.8em, left=2em of box9] (box6) {};
	\foreach \x in {1,...,4}
	{\draw node[head](e6\x)[yshift=0.5em, xshift=(\x-2)*1.1em-0.5em, below=0mm of box6.center]{};}
	
	\draw node[convBox][left=2em of box9] (box5) {};
	\foreach \x in {1,...,4}
	{\draw node[relation](e5\x)[yshift=0.5em, xshift=(\x-2)*1.1em-0.5em, below=0mm of box5.center]{};}
	
	\draw node[convBox][yshift=-3.8em, left=2em of box9] (box4) {};
	\foreach \x in {1,...,4}
	{\draw node[tail](e4\x)[yshift=0.5em, xshift=(\x-2)*1.1em-0.5em, below=0mm of box4.center]{};}

	\draw node[convolution][left=2em of box5] (filterBox1) {};
	\foreach \x in {1,...,3}
	{\draw node[conv, fill=amber](filter1\x)[yshift=0.5em, xshift=(\x-2)*1.1em, below=0mm of filterBox1.center]{};}
	\foreach \x in {1,...,3}
	{\draw node[conv, fill=azure](filter2\x)[yshift=0.5em, xshift=(\x-2)*1.1em+0.2em, below=0mm of filterBox1.center]{};}
	\foreach \x in {1,...,3}
	{\draw node[conv, fill=awesome](filter3\x)[yshift=0.5em, xshift=(\x-2)*1.1em+0.4em, below=0mm of filterBox1.center]{};}
	
	\draw node[embBox][minimum width=1.3em, minimum height=5em, xshift=-1.5em, left=4em of filterBox1] (box1) {};
	\foreach \x in {1,...,4}
	{\draw node[head, fill=amber!30](e1\x)[yshift=(\x-2)*1.1em, below=0mm of box1.center]{};}
	\draw node[title][above=0mm of box1] (headText) {\small $\text{h}$};
	
	\draw node[embBox][minimum width=1.3em, minimum height=5em, left=4em of filterBox1] (box2) {};
	\foreach \x in {1,...,4}
	{\draw node[relation, fill=azure!30](e2\x)[yshift=(\x-2)*1.1em, below=0mm of box2.center]{};}
	\draw node[title][above=0mm of box2] (relationText) {\small $\text{r}$};
	
	\draw node[embBox][minimum width=1.3em, minimum height=5em, xshift=1.5em,left=4em of filterBox1] (box3) {};
	\draw node[title][above=0mm of filterBox1] (filterTitle) {\small $3.\omega^{\text{1} \times \text{3}}$};
	\foreach \x in {1,...,4}
	{\draw node[tail, fill=awesome!30](e3\x)[yshift=(\x-2)*1.1em, below=0mm of box3.center]{};}
	\draw node[title][above=0mm of box3] (tailText) {\small $\text{t}$};
	
	\draw node[convLayer, color=red][yshift=1.71em, left=2em of filterBox1] (conv1){};
	\draw node[convLayer, color=green][yshift=0.61em, left=2em of filterBox1] (conv2){};
	\draw node[convLayer, color=blue][yshift=-0.5em, left=2em of filterBox1] (conv3){};
	\draw node[convLayer, color=amber][yshift=-1.6em, left=2em of filterBox1] (conv4){};
	
	\foreach \x in {1,...,4}
	\draw [arrowStrong] (conv\x.east) -- (filterBox1.west);
	
	\foreach \x in {4,...,6}
	\draw [arrowStrong] (filterBox1.east) -- (box\x.west);
	
	\draw [arrow] (box4.east) -- ([xshift=0.2mm, yshift=0.5mm]box9.south west);
	\draw [arrow] (box5.east) -- (box9.west);
	\draw [arrow] (box6.east) -- ([xshift=0.2mm, yshift=-0.5mm] box9.north west);
	
	\foreach \x in {1,...,4}
	\draw [arrow] (e9\x.east) -- (lossConvKB.west);
	\foreach \x in {1,...,4}
	\draw [arrow] (e8\x.east) -- (lossConvKB.west);
	\foreach \x in {1,...,4}
	\draw [arrow] (e7\x.east) -- (lossConvKB.west);
	
\end{tikzpicture}
	}
	\caption{Illustration of the decoder layers of the ConvKB model with 3 filters}
	\label{fig:decoderConvKB}
\end{figure}

After mapping entities and relations into a low-dimensional space, the model employs ConvKB \cite{nguyen2017novel} to analyze the global features of a triple $t_{ijk}$ across each dimension, thereby generalizing transformation patterns through convolutional layers. The scoring function with the learned feature mappings is defined as follows:

\begin{equation}
	f(t_{ijk}) = \Big( \bigparallel_{m=1}^{\Omega} \text{ReLU} ([\overrightarrow{e_i}, \overrightarrow{r_k}, \overrightarrow{h_j}] \ast \omega^m)\Big).\mathbf{W}
\end{equation}

where $\omega^m$ denotes the $m$-th convolutional filter,  
$\Omega$ is the hyperparameter representing the number of convolutional layers,  
$\ast$ denotes the convolution operation, and  
$\mathbf{W} \in \mathbb{R}^{\Omega k \times 1}$ is the linear transformation matrix used to compute the final score for the triple.  

The model is trained using a soft-margin loss function as follows:

\begin{equation}
	\label{eq:lossConvKB}
	\mathcal{L} = \sum_{t^k_{ij} \in \{S \cup S'\}} \text{log}(1 + \exp(l_{t^k_{ij}} \cdot f(t^k_{ij}))) + \frac{\lambda}{2} \parallel{\mathbf{W}}\parallel_2^2
\end{equation}

where 

\[
l_{t^k_{ij}} = 
\begin{cases}
	1 & \text{for } t^k_{ij} \in S \\
	-1 & \text{for } t^k_{ij} \in S' \\
\end{cases}
\]

The final output of the ConvKB model is the ranking score corresponding to each prediction.

\section{Experiment}
\label{sec:experiment}

\begin{figure}[h]
	\centering
	\label{fig:dataset}
	\resizebox{\linewidth}{!}{
	\begin{tikzpicture}
		[every axis/.style={
			ybar,
			scale only axis,
			ymin=0, ymax= 130000,
			width=0.5\textwidth,
			height=0.4\textwidth,
			legend style={at={(20em,5em)}, anchor=east},
			bar width=1em,
			scaled y ticks=false,
			xtick=data,
			font=\scriptsize,
			symbolic x coords={FB15k,FB15k-237,WN18,WN18RR,YAGO3-10},
			nodes near coords,
			nodes near coords align={vertical},
		}]
		\pgfplotsset{
			compat=newest,
			major grid style=blue,
			xlabel near ticks,
			ylabel near ticks
		}
		
		\begin{axis}[]
			\addplot [fill=awesome] coordinates {
				(FB15k,14951)
				(FB15k-237,14541)
				(WN18,40943)
				(WN18RR,40559)
				(YAGO3-10,123182)
			};
			\addplot [fill=azure] coordinates {
				(FB15k,1345)
				(FB15k-237,237)
				(WN18,18)
				(WN18RR,11)
				(YAGO3-10,37)
			};
			\legend{Entities, Relations}
		\end{axis}
	\end{tikzpicture}
}
	
\end{figure}

In this section, we outline the datasets used for empirical evaluation and assess the performance of our two proposed methods for incorporating new knowledge into the knowledge graph. To this end, we simulate the addition of new information by treating the test set as a batch of unseen triples and re-evaluate model performance using the validation set. The experimental results are summarized in \autoref{tab:resultOnFreeBase} and \autoref{tab:resultOnWordNet}.

\begin{table}[h]
	\begin{center}
				\resizebox{\linewidth}{!}{%
			\begin{tabular}{llllll}
				\hline
				&          &           & \multicolumn{3}{l}{\# Edges}    \\ \cline{4-6}
				
				Dataset   & Entities & Relations & Training & Validation & Test    \\ \hline
				FB15k     & 14,951   & 1,345     & 483,142  & 50,000     & 59,071 \\
				FB15k-237 & 14,541   & 237       & 272,115  & 17,535     & 20,466  \\
				WN18      & 40,943   & 18        & 141,442  & 5,000      & 5,000   \\
				WN18RR    & 40,559   & 11        & 86,835   & 3,034       & 3,134    \\
				YAGO3-10    & 123,182   & 37        & 1,079,040   & 5,000       & 5,000  \\
				\hline
			\end{tabular}
					}
		\caption{Dataset Information}
		\label{tab:datasetInfo}
	\end{center}
\end{table}

\subsection{Training Datasets}
\label{sec:DataTraining}

\begin{figure}[h]
	\centering
	\label{fig:dataset_split}
	
	\pgfplotstableread[row sep=\\,col sep=&]{
		Dataset & Entities & Relations & Training & Validation & Test\\
		FB15k & 14951 & 1345 & 483142 & 50000 & 59071  \\
		FB15k-237 & 14541 & 237 & 272115 & 17535 & 20466  \\
		WN18 & 40943 & 18 & 141442 & 5000 & 5000  \\
		WN18RR & 40559 & 11 & 86835 & 3034 & 3134  \\
		YAGO3-10 & 123182 & 37 & 1079040 & 5000 & 5000  \\
	}\mydata
	\resizebox{\linewidth}{!}{%
		\begin{tikzpicture}
			\begin{axis}[
				xbar stacked,
				tick align = outside, xtick pos = left,
				scale only axis,
				scaled x ticks=false,
				every node near coord/.style={/pgf/number format/fixed},
				xticklabel style={/pgf/number format/fixed},
				width=\textwidth,
				height=0.3\textwidth,
				font=\scriptsize,
				legend style={at={(0.5,-0.15)}, anchor=north, legend columns=-1},
				bar width=1.5em,
				ytick=data,
				y dir = reverse,
				yticklabels from table={\mydata}{Dataset},
				]
				\addplot[fill=azure] table [y expr=\coordindex,x=Training]{\mydata};
				\addplot+[fill=awesome] table [y expr=\coordindex,x=Validation]{\mydata};
				\addplot+[fill=amber,
				point meta=x,
				nodes near coords = {\pgfmathprintnumber[precision=1]{\pgfplotspointmeta}},
				nodes near coords align={anchor=west},
				every node near coord/.append style={
					black,
					fill=white,
					fill opacity=0.75,
					text opacity=1,
					outer sep=\pgflinewidth
				}] table [y expr=\coordindex,x=Test]{\mydata};
				\legend{Training, Validation, Test};
			\end{axis}
		\end{tikzpicture}
	}
\end{figure}

In our experiments, we evaluate our approach on four widely used benchmark datasets: FB15k, FB15k-237 (\cite{toutanova2015observed}), WN18, and WN18RR (\cite{dettmers2018convolutional}). Each dataset is divided into three subsets: training, validation, and test sets. Detailed statistics for these datasets are presented in \autoref{tab:datasetInfo}. 

Each dataset consists of a collection of triples in the form \(\langle head, relation, tail \rangle\). FB15k and WN18 are derived from the larger knowledge bases FreeBase and WordNet, respectively. However, they contain a large number of inverse relations, which allow most triples to be easily inferred. To address this issue and to better reflect real-world link prediction scenarios, FB15k-237 and WN18RR were constructed by removing such inverse relations.

\subsubsection{FB15k Dataset}

This dataset was constructed by the research group of A. Bordes and N. Usunier \cite{bordes2013translating} by extracting data from the Wikilinks database\footnote{https://code.google.com/archive/p/wiki-links/}. The Wikilinks database contains hyperlinks to Wikipedia articles, comprising over 40 million mentions of approximately 3 million entities. The authors extracted all facts related to a given entity that is mentioned at least 100 times across different documents, along with all facts associated with those entities (including child entities mentioned in the corresponding Wikipedia articles), excluding attributes such as dates, proper nouns, etc. 

Entities with degree \(n\) were normalized by converting multi-way edges into sets of binary relations—enumerating all edges and relations between each pair of nodes.


\subsubsection{FB15k-237 Dataset}

This dataset is a subset of FB15k, constructed by Toutanova and Chen \cite{toutanova2015observed}, motivated by the observation that FB15k suffers from test leakage, where models are exposed to test facts during training. This issue arises due to the presence of duplicate or inverse relations in FB15k. FB15k-237 was created to provide a more challenging benchmark. The authors selected facts related to the 401 most frequent relations and eliminated all redundant or inverse relations. Additionally, they ensured that no entities connected in the training set were directly linked in the test and validation sets.


\subsubsection{WN18 Dataset}

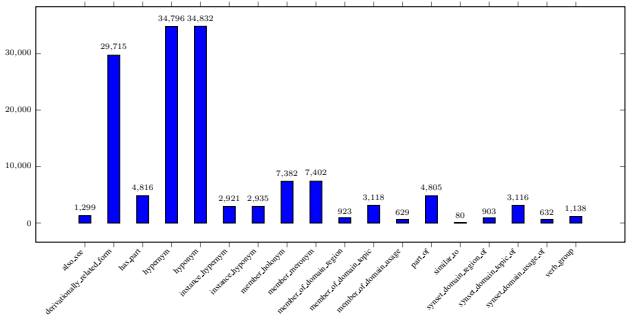
\begin{center}
	\resizebox{\linewidth}{!}{%
		\begin{tikzpicture}
			[every axis/.style={
				ybar,
				scale only axis,
				width=\textwidth,
				height=0.4\textwidth,
				xtick=data,
				x tick label style={rotate=45, anchor=east},
				legend style={at={(20em,5em)}, anchor=east},
				bar width=1em,
				scaled y ticks=false,
				font=\scriptsize,
				symbolic x coords={
					also\_see,
					derivationally\_related\_form,
					has\_part,
					hypernym,
					hyponym,
					instance\_hypernym,
					instance\_hyponym,
					member\_holonym,
					member\_meronym,
					member\_of\_domain\_region,
					member\_of\_domain\_topic,
					member\_of\_domain\_usage,
					part\_of,
					similar\_to,
					synset\_domain\_region\_of,
					synset\_domain\_topic\_of,
					synset\_domain\_usage\_of,
					verb\_group},
				nodes near coords,
				nodes near coords align={vertical},
			}]
			\pgfplotsset{
				compat=newest,
				major grid style=blue,
				xlabel near ticks,
				ylabel near ticks
			}
			
			\begin{axis}[]
				\addplot [fill=blue] coordinates {
					(also\_see,1299)
					(derivationally\_related\_form,29715)
					(has\_part,4816)
					(hypernym,34796)
					(hyponym,34832)
					(instance\_hypernym,2921)
					(instance\_hyponym,2935)
					(member\_holonym,7382)
					(member\_meronym,7402)
					(member\_of\_domain\_region,923)
					(member\_of\_domain\_topic,3118)
					(member\_of\_domain\_usage,629)
					(part\_of,4805)
					(similar\_to,80)
					(synset\_domain\_region\_of,903)
					(synset\_domain\_topic\_of,3116)
					(synset\_domain\_usage\_of,632)
					(verb\_group,1138)
				};
			\end{axis}
		\end{tikzpicture}
	}
\end{center}

This dataset was introduced by the authors of TransE \cite{bordes2013translating}, and is extracted from WordNet\footnote{https://wordnet.princeton.edu/}, a lexical knowledge graph ontology designed to provide a dictionary/thesaurus to support NLP tasks and automated text analysis. In WordNet, entities correspond to synsets (i.e., \textit{word senses}), and relations represent lexical connections among them (e.g., “hypernym”). 

To construct WN18, the authors used WordNet as a starting point and iteratively filtered out entities and relations that were infrequently mentioned.


\subsubsection{WN18RR Dataset}

This dataset is a subset of WN18, constructed by DeŠmers et al.\cite{dettmers2017convolutional}, who also addressed the issue of test leakage in WN18. To tackle this issue, they created the WN18RR dataset, which is significantly more challenging, by applying a similar methodology to that used in FB15k-237 \cite{toutanova2015observed}.

\subsection{Evaluation Metrics}

In this section, we describe the evaluation metrics, experimental environment, and datasets used to assess the proposed method. These metrics are widely adopted for evaluating link prediction models on knowledge graphs. We compare our approach against four other state-of-the-art methods reported in \cite{rossi2020knowledge}.

\textbf{Hits@K (H@K)}

This metric measures the proportion of correct predictions whose rank is less than or equal to the threshold \(K\):
\[
H@K = \frac{\mid \{ q \in Q : \text{rank}(q) \leq K \} \mid}{\mid Q \mid}
\]

\textbf{Mean Rank (MR)}

This metric calculates the average rank of the correct entity in the prediction. A lower value indicates better model performance:
\[
MR = \frac{1}{\mid Q \mid} \sum_{q \in Q} \text{rank}(q)
\]
Here, \(\mid Q \mid\) denotes the total number of queries, which equals the size of the test or validation set. During evaluation, we perform both head and tail entity predictions for each triple. For example, we predict both \(\langle ?,~ \text{relation},~ \text{tail} \rangle\) and \(\langle \text{head},~ \text{relation},~ ? \rangle\). The variable \(q\) denotes a query, and \(\text{rank}(q)\) indicates the rank position of the correct entity. The final MR score is the average rank over all head and tail predictions.

Clearly, this metric ranges from \([1,~|\text{number of entities}|]\), as a node can connect to at most \(n-1\) other nodes plus a self-loop. However, this metric is highly sensitive to outliers, as certain relations may yield extremely low rankings for correct entities. To address this issue, our method—as well as other recent works—also adopts the Mean Reciprocal Rank (MRR) metric.

\textbf{Mean Reciprocal Rank (MMR)}
This is the Mean Reciprocal Rank (MRR), calculated as the reciprocal of the average rank obtained for a correct prediction. Higher values indicate better model performance. Since this metric takes the reciprocal of each rank, it helps mitigate the noise sensitivity encountered in the Mean Rank (MR) metric:
\[
MRR =\frac{1}{\mid Q \mid} \sum_{q \in Q} \frac{1}{\text{rank}(q)}
\]

\subsection{Training Methodology}

\subsubsection{Training with the GCAT Model}

We first initialize the entity and relation embeddings using the TransE model \cite{bordes2013translating}. To construct negative (invalid) triples, we randomly replace either the head or tail entity in a valid triple with another entity sampled from the entity set.

The training process is divided into two phases. The first phase serves as an encoder, transforming the initialized embeddings into new embeddings that aggregate neighborhood information using the GCAT model. This produces updated embeddings for both entities and relations. The second phase serves as a decoder, performing link prediction by incorporating $n$-hop neighborhood information. This enables the model to better aggregate context from neighboring entities. Furthermore, we incorporate auxiliary relations to enrich the local structure in sparse graphs.

We use the Adam optimizer with a learning rate of $\mu = 0.001$. The final embedding dimension for both entities and relations is set to 200. 

\subsection{Experimental Results}
\label{sec:Experiment}

As previously mentioned, our rule-based model can be fully executed on a standard laptop. In our experiment, the machine configuration was as follows: T480, Core i5 8th Gen, 16GB RAM, 4 cores and 8 threads. The source code was implemented in Python version 3.6, utilizing only built-in Python functions without any third-party libraries. The experiments were conducted on four widely-used datasets: FB15k, FB15-237, WN18, and WN18RR. Detailed information about these datasets is provided in \autoref{sec:DataTraining} under the training datasets section.

\begin{table}[h]
	\begin{center}
		\caption{Experimental results on the FB15k and FB15k-237 datasets}
		\label{tab:resultOnFreeBase}%
		\resizebox{\linewidth}{!}{%
			\begin{tabular}{l|l|l|l|l|l|l|l|l|}
				\cline{2-9}
				& \multicolumn{4}{c|}{\textbf{FB15k}}                   & \multicolumn{4}{c|}{\textbf{FB15k-237}}                   \\ \cline{2-9} 
				& \textbf{H@1} & \textbf{H@10} & \textbf{MR} & \textbf{MRR} & \textbf{H@1} & \textbf{H@10} & \textbf{MR} & \textbf{MRR} \\ \hline
				\multicolumn{1}{|l|}{ComplEx} & 81.56        & 90.53         & 34          & 0.848        & 25.72        & 52.97         & 202        & 0.349        \\ \hline
				\multicolumn{1}{|l|}{TuckER}  & 72.89        & 88.88         & 39          & 0.788        & 25.90        & 53.61         & 162         & 0.352        \\ \hline
				\multicolumn{1}{|l|}{TransE}  & 49.36        & 84.73         & 45          & 0.628        & 21.72        & 49.65         & 209         & 0.31        \\ \hline
				\multicolumn{1}{|l|}{RoteE}   & 73.93        & 88.10         & 42          & 0.791        & 23.83        & 53.06         & 178         & 0.336        \\ \hline
				\multicolumn{1}{|l|}{ConvKB}  & 59.46        & 84.94         & 51         & 0.688        & 21.90        & 47.62         & 281         &0.305        \\ \hline
				\multicolumn{1}{|l|}{\textbf{GCAT}}     &  70.08            &     91.64    &  38    &   0.784    & 36.06     &    58.32   &  211  &    0.4353  \\ \hline
			\end{tabular}
		}
	\end{center}
\end{table}

\begin{table}[h]
	\begin{center}
		\caption{Experimental results on the WN18 and WN18RR datasets}
		\label{tab:resultOnWordNet}%
		\resizebox{\linewidth}{!}{%
			\begin{tabular}{l|l|l|l|l|l|l|l|l|}
				\cline{2-9}
				& \multicolumn{4}{c|}{\textbf{WN18}}                              & \multicolumn{4}{c|}{\textbf{WN18RR}}                            \\ \cline{2-9} 
				& \textbf{H@1}   & \textbf{H@10}  & \textbf{MR}  & \textbf{MRR}   & \textbf{H@1}   & \textbf{H@10}  & \textbf{MR}  & \textbf{MRR}   \\ \hline
				\multicolumn{1}{|l|}{ComplEx} & 94.53          & 95.50          & 3623         & 0.349          & 42.55          & 52.12          & 4909         & 0.458          \\ \hline
				\multicolumn{1}{|l|}{TuckER}  & 94.64          & 95.80          & 510          & 0.951          & 42.95          & 51.40          & 6239         & 0.459          \\ \hline
				\multicolumn{1}{|l|}{TransE}  & 40.56          & 94.87          & 279          & 0.646          & 2.79           & 94.87          & 279          & 0.646          \\ \hline
				\multicolumn{1}{|l|}{RoteE}   & 94.30          & 96.02          & 274          & 0.949          & 42.60          & 57.35          & 3318         & 0.475          \\ \hline
				\multicolumn{1}{|l|}{ConvKB}  & 93.89          & 95.68          & 413          & 0.945          & 38.99          & 50.75          & 4944         & 0.427          \\ \hline
				\multicolumn{1}{|l|}{\textbf{GCAT}}     &                &        &        &                &       35.12         &        57.01         &      \underline{1974}       &  0.4301           \\ \hline
			\end{tabular}
		}
	\end{center}
\end{table}

\section{Conclusion}
\label{chap:Conclusion}

In this study, we proposed GCAT (Graph Collaborative Attention Network), a deep learning-based architecture designed for link prediction in knowledge graphs. Through extensive experiments across multiple benchmark datasets, GCAT consistently outperformed rule-based methods, particularly on complex and filtered datasets such as FB15k-237 and WN18RR. These results highlight GCAT’s superior ability to capture semantic nuances and relational diversity by leveraging attention-based collaborative message passing.

While rule-based approaches offer advantages in interpretability and reduced training costs, their performance heavily depends on the presence of repetitive or inverse relational patterns. When such patterns are filtered out, as in more challenging benchmarks, their effectiveness drops significantly. In contrast, GCAT remains robust by embedding entities and relations into expressive latent spaces, allowing for more accurate generalization across diverse graph structures.

Furthermore, the integration mechanisms introduced in GCAT for dynamic knowledge updates outperform traditional baselines, demonstrating strong adaptability in evolving graph environments. These contributions position GCAT as a state-of-the-art solution that not only addresses the limitations of symbolic reasoning but also scales effectively to real-world, temporally rich knowledge graphs.

In future work, we plan to explore heterogeneous embedding dimensionality for entities and relations to further enhance model expressiveness. Additionally, integrating temporal attention mechanisms remains a promising direction to capture the dynamic nature of factual knowledge in time-evolving graph structures.

{
    \small
    \bibliographystyle{ieeenat_fullname}
    \bibliography{reference}
}


\end{document}